%% file: main.tex
\pdfoutput=1
\documentclass[11pt]{scaleai-paper}

\usepackage{amsmath}
\usepackage{mathtools}  
\usepackage{amsfonts}
\usepackage{amssymb}
\usepackage{amsthm}
\usepackage{booktabs}
\usepackage{tabularx}
\usepackage{tabulary}
\usepackage{multirow}
\usepackage{subcaption}
\usepackage{float}
\usepackage{wrapfig}
\usepackage[square,numbers,sort&compress]{natbib}
\usepackage{xspace}
\usepackage{url}
\usepackage{enumitem}
\usepackage{tikz}
\usetikzlibrary{positioning,fit,backgrounds,calc,svg.path}
\usepackage[colorlinks=true,linkcolor=scaleLink,citecolor=scaleLink,urlcolor=scaleLink]{hyperref}
\usepackage{fontawesome5}  
\usepackage[capitalise,nameinlink]{cleveref}
\AtBeginDocument{
  }
\crefname{section}{\S}{\S\S}\Crefname{section}{\S}{\S\S}
\crefname{subsection}{\S}{\S\S}\Crefname{subsection}{\S}{\S\S}
\crefname{appendix}{Appendix}{Appendices}\Crefname{appendix}{Appendix}{Appendices}
\crefname{observation}{Observation}{Observations}
\Crefname{observation}{Observation}{Observations}


\providecolor{scaleAmber}{HTML}{E09A3B}
\providecolor{scaleCoral}{HTML}{D95358}
\providecolor{scalePurple}{HTML}{A37CAD}
\providecolor{scaleSky}{HTML}{6FCBEF}
\providecolor{scaleTerracotta}{HTML}{E27553}
\newif\ifscalebuild \scalebuildtrue  
\newcommand{\dosewrapwidth}{0.40}\newcommand{\dosewrapbottom}{-0.6}\newcommand{\dosefigwidth}{0.38}


\newif\ificlrstatements \iclrstatementstrue
\newif\ifdraftnotes \draftnotesfalse  
\ifdraftnotes\usepackage{marginnote}\fi  
\usepackage{tcolorbox}
\usepackage{needspace}
\newcommand{\reviewmargin}[2]{\ifdraftnotes\marginnote{\footnotesize\raggedright\textcolor{#1}{[#2]}}\fi}

\newcommand{\clnote}[1]{\reviewmargin{scaleCoral}{claude: #1}}

\providecolor{scaleGray}{HTML}{7D7D7D}
\newif\ifagentmarks \agentmarkstrue
\newcommand{\agentedit}[1]{\ifagentmarks{\color{scaleGray}#1}\else#1\fi}

\usepackage[normalem]{ulem}

\newtheorem{proposition}{Proposition}
\newtheorem{observation}{Observation}
\newcolumntype{Y}{>{\RaggedRight\arraybackslash}X}
\let\svthefootnote\thefootnote
\newcommand\freefootnote[1]{%
  \let\thefootnote\relax%
  \footnotetext{#1}%
  \let\thefootnote\svthefootnote%
}

\newcommand{\emailicon}{\raisebox{-0.12em}{\resizebox{!}{0.78em}{%
  \tikz[yscale=-1]\fill svg {M48 64C21.5 64 0 85.5 0 112c0 15.1 7.1 29.3 19.2 38.4L236.8 313.6c11.4 8.5 27 8.5 38.4 0L492.8 150.4c12.1-9.1 19.2-23.3 19.2-38.4c0-26.5-21.5-48-48-48L48 64zM0 176L0 384c0 35.3 28.7 64 64 64l384 0c35.3 0 64-28.7 64-64l0-208L294.4 339.2c-22.8 17.1-54 17.1-76.8 0L0 176z};}}}

\newcommand{\gpt}{\texttt{gpt-4o-mini}}
\newcommand{\claude}{\texttt{claude-sonnet-4-6}}

\papertype{Scale AI Research}
\contact{\emailicon\ \texttt{minglai.yang@scale.com}}

\title{Rubric Dropout: A Simple Way to Mitigate Reward Hacking in Rubric-Based RL}

\author[1,$\dagger$]{Minglai Yang}
\author[2]{Xinyu Guo}
\author[3]{Zhengliang Shi}
\author[1]{Utkarsh Tyagi}
\author[1,4]{Mian Zhang}
\author[1]{\mbox{Razvan-Gabriel Dumitru}}  
\author[1]{Sunjie Hou}
\author[1]{Yunzhong He}
\author[1]{Daniel Yue Zhang}
\author[5]{Dan Roth}
\author[1]{Ying Liu}
\affil[1]{Scale AI}
\affil[2]{University of Arizona}
\affil[3]{Carnegie Mellon University}
\affil[4]{University of Texas at Dallas}
\affil[5]{University of Pennsylvania}

\makeatletter\g@addto@macro\normalsize{%
  \setlength\abovedisplayskip{4pt plus 1pt minus 2pt}\setlength\belowdisplayskip{4pt plus 1pt minus 2pt}%
  \setlength\abovedisplayshortskip{0pt plus 1pt}\setlength\belowdisplayshortskip{3pt plus 1pt minus 1pt}}\makeatother
\begin{document}

\freefootnote{${}^\dagger$Corresponding author: Minglai Yang (\emailicon\ \texttt{minglai.yang@scale.com}).}

\maketitle

\input{sections/00_abstract}

\input{sections/01_intro}

\input{sections/02_related}

\input{sections/03_method}

\input{sections/04_experiments}

\input{sections/05_ablations}

\input{sections/06_discussion}

\input{sections/07_conclusion}

\input{sections/08_limitations}


\bibliographystyle{abbrvnat}
\bibliography{references}

\appendix
\crefalias{section}{appendix}\crefalias{subsection}{appendix}

\input{sections/09_appendix}

\end{document}

%% file: sections/00_abstract.tex
\begin{abstract}
Rubric-based reinforcement learning, in which an LLM judge scores responses against prompt-specific criteria, has become a standard way to post-train language models on tasks without verifiable answers.
The rubric, however, is only a proxy for response quality, and a policy trained against it long enough learns to exploit the difference.
We train models with GRPO on medical and science rubrics and grade their responses on out-of-distribution (OOD) benchmarks throughout training with two judges: the training judge and a stronger gold judge.
The two scores first rise together and then diverge.
The policy is learning to satisfy the training judge without writing better responses, which is a form of reward hacking. To mitigate it, we propose \textbf{\textit{Rubric Dropout}}, which applies dropout to rubric criteria: at each training step, a random subset of each rubric's criteria is removed from the reward, so the rewarded subset of criteria is resampled at each step.
All responses sampled for the same prompt are scored on the same subset, so their rewards remain comparable, and evaluation always uses the full rubric.
We also introduce \textbf{\textit{Block Dropout}}, which drops semantically similar criteria together so that a near-duplicate of a dropped criterion cannot keep rewarding the same behavior.
Across both benchmarks and models from 1.7B to 8B, dropout improves the gold judge's score on OOD benchmarks by up to 3\% on HealthBench-Hard and 7\% on ResearchQA, and it reduces both the gap between the two judges and over-crediting. It keeps in-domain reward close to the baseline, and dropout rates of 30--50\% work best.
Block Dropout improves the gold judge's OOD score further at the same dropout rate.
Changing which criteria are rewarded at each step is a simple, inexpensive way to make rubric rewards harder to exploit.
\end{abstract}

%% file: sections/01_intro.tex
\section{Introduction}


\begin{figure}[!t]
 \centering
 \includegraphics[width=0.97\linewidth]{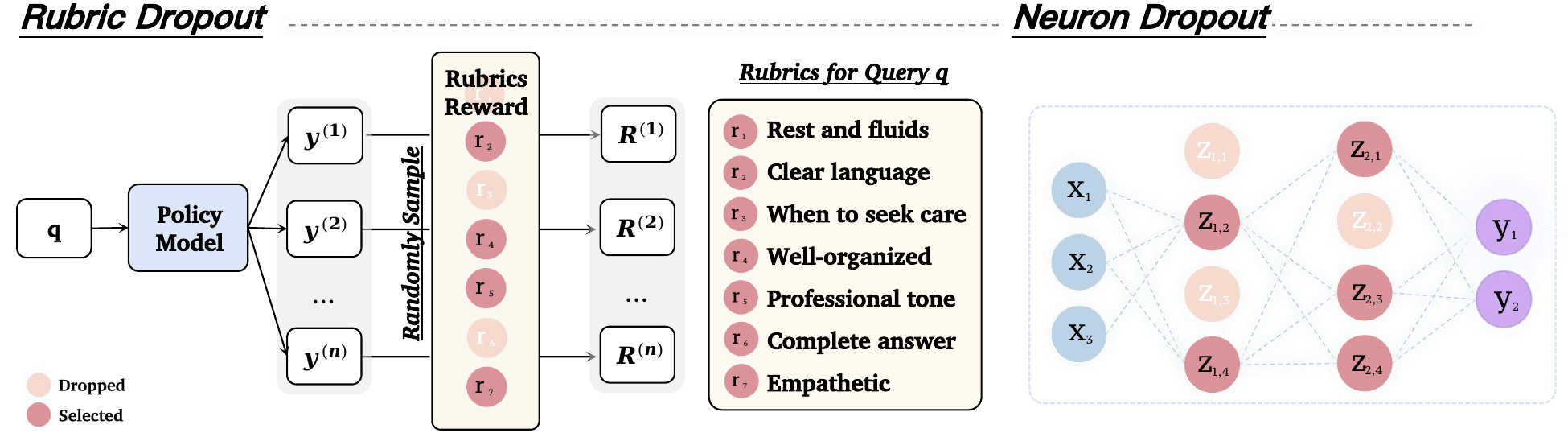}
 \caption{
\textbf{Rubric Dropout randomly drops rubric criteria from the training reward.} Left: every rollout of a query is scored on the same randomly sampled sub-rubric (faded criteria are dropped). The mask is re-drawn at each training step. Dropout is train-only and evaluation uses the full rubric.}
 \label{fig:teaser}
 \vspace{-10pt}
\end{figure}


Reinforcement learning with verifiable rewards (RLVR) works well in domains where there is a ground truth to check the answer against \citep{lambert2024tulu3,
deepseekai2025r1}. However, many of the tasks we most want large language models to be
good at, such as giving medical advice \citep{arora2025healthbench,zhang-etal-2025-llmeval} and explaining a research area, are open-ended, with no ground truth. A common approach is
\emph{Rubric-based} RL: write down a list of criteria for each prompt, have an LLM judge grade each criterion, and use the weighted fraction satisfied as the reward
\citep{gunjal2025rubrics, huang2025rubric}. The recipe is attractive because rubrics make quality explicit and auditable, and recent work reports strong gains from it.

The recipe also has a built-in weakness.
The rubric score the training judge assigns is only a \emph{proxy} for quality, and this proxy is \emph{fixed}.\clnote{2026-09-25: ``proxy'' now means one thing throughout: the training judge's rubric score.} The same criteria are scored at
every training step, many of them generic templates that repeat across prompts
(``uses clear language'', ``well-organized'').
Prompt-specific, verifiable criteria would be ideal. At dataset scale, though, keeping every criterion that specific and verifiable is hard, and generic templates end up in the training set.
This immutability makes the proxy exploitable. A criterion rewarded identically at every step is a stable target. Once the policy finds a cheap, surface-level way to satisfy it, the shortcut is reinforced at every subsequent step, and when the criterion is a
shared template, it is reinforced on every prompt at once. A policy that learns to open every answer with a tidy bulleted summary satisfies ``well-organized'' everywhere, regardless of content.
Classic results on reward
misspecification indicate that optimizing hard against a fixed, imperfect proxy ends
in reward hacking \citep{amodei2016concrete, skalse2022reward,
pan2022misspecification}, and \citet{gao2023overoptimization} showed exactly
this for learned reward models. 

To see how severe this is for rubric rewards in practice, we trained with the
standard recipe and tracked quality with a signal the training loop cannot see.
Every 20 steps we graded an OOD evaluation set with two judges: the training judge, whose score is the \emph{proxy}, and a stronger cross-family judge, whose score is the \emph{gold}. We train against the proxy, not the gold judge, for the same reason practitioners do. The proxy is cheap enough to grade every rollout at every step, while the gold judge costs $20\times$ as much\footnote{Per million input/output tokens: \gpt{} \$0.15/\$0.60, \claude{} \$3/\$15.} and is affordable only for periodic evaluation. On two independent train$\to$eval pairs, RubricHub-Medical$\to$HealthBench-Hard and RubricHub-Science$\to$ResearchQA, the proxy and gold scores diverge. The gold score rises, peaks, and then declines, while the proxy score keeps improving and the set of criteria it overclaims keeps growing. The proxy$-$gold gap widens from 29\% to as much as 44\% on HealthBench-Hard, and on ResearchQA gold falls 22 points from its peak. We read this divergence as a sign of hacking rather than judge bias. A merely biased judge would shift the gold curve by a constant and would not bend it downward while the proxy rises. 
This test alone cannot rule out a gold judge whose errors change with the policy. We therefore also look at which criteria the proxy overclaims (\cref{sec:where-gain}) and at how the base policy's outputs change (\cref{sec:discussion}); both point to template-like shortcuts rather than to a bias in the gold judge.
These runs indicate that rubric rewards are hacked in practice. Our study has two objectives: 1) Quantify reward hacking as the gap between training reward and a quality signal independent of the training judge and rubrics.
2) Develop a dynamic rubric-reward method that reduces reward hacking and works with existing RL infrastructure.

First, we introduce a two-judge evaluation protocol: at regular checkpoints, the \textit{proxy} judge and a stronger \textit{gold} judge grade the same responses on OOD tasks. Because hacking sets in late in training, this \agentedit{in-loop} view catches degradation that end-of-run scores miss.

For the second objective, we propose \textit{\textbf{Rubric Dropout}} (\cref{fig:teaser}). At every
training step, we randomly drop a fraction of the rubric's criteria before
computing the reward, so the policy is scored on a sampled sub-rubric each time.
No single criterion is then guaranteed to be rewarded at every step. This means the policy cannot easily exploit the reward with simple template-based strategies.
The change is one line in the reward function, has a single hyperparameter, and
adds no judge calls. We draw one mask per rollout group,
so all rollouts of a prompt are graded on the same sub-rubric.
We prove that under this scheme the advantage does not depend on the choice of reward normalizer. We also show that, before group standardization, dropout only rescales the expected advantage, while it acts as a variance regularizer on updates that depend heavily on any single criterion.
Because near-duplicate criteria can preserve a signal even when one of them is dropped, we also introduce \textit{\textbf{Block Dropout}}, which jointly drops semantically similar criteria (\cref{sec:blockmethod}).

Using this measurement and method, we compare three conditions throughout the paper: no dropout (\emph{base}), dropout at $f{\approx}30\%$, and dropout at $f{\approx}50\%$. In both domains and at three model sizes, dropout mitigates the hacking. Peak scores are tied between base and dropout, and there is no measurable in-domain cost. We also find that dropout preserves output diversity late in training, while the base policy collapses to low-entropy, template-like responses (\cref{fig:entropy}). Sweeping $f$ shows a broad plateau, and reweighting, used in place of dropout, scores \emph{below} base.

Our contributions are summarized as follows:

\begin{enumerate}[
  leftmargin=*,
  itemsep=1pt,
  parsep=0pt,
  topsep=0pt,
  partopsep=0pt
]
  \item We quantify OOD reward hacking by evaluating the policy every
  20 training steps with both the training judge and a stronger
  cross-family judge. On two independent benchmark pairs, the
  training-judge score continues to rise after the stronger-judge score
  peaks.

  \item We propose Rubric Dropout, a simple regularizer for mitigating
  rubric reward hacking. Its group-shared masking scheme is theoretically
  compatible with GRPO. We further introduce block dropout, which drops
  semantically correlated criteria together.

  \item
Across two benchmark pairs and models from 1.7B to 8B, Rubric Dropout improves OOD gold reward and reduces reward hacking, with no loss of in-domain performance.
\end{enumerate}



%% file: sections/02_related.tex
\section{Related Work}
\label{sec:background}

\paragraph{Rubric-based reinforcement learning.}
Grading free-form text against explicit criteria began as an evaluation idea \citep{zheng2023mtbench,
arora2025healthbench} and has since been adopted for training.
\mbox{QA-LIGN} grades responses against question checks decomposed from a constitution \citep{dineen2025qalign}. RaR \citep{gunjal2025rubrics} and Rubric Anchors \citep{huang2025rubric} use rubric scores directly as RL rewards. Checklist feedback does the same with per-instruction checklists \citep{checklists}, and follow-up work scales rubric generation \citep{openrubrics} or uses rubrics to scaffold exploration \citep{ruscarl}. These methods all optimize against a fixed rubric per prompt and evaluate in-domain quality. Recent work addresses the brittleness of a fixed rubric by changing its \emph{content}. RIFL, for example, appends a fixed
set of negative rubrics that penalize known failure modes \citep{advancedif}.
Writing these negative rubrics adds authoring cost. We avoid this cost by keeping the rubric exactly as written and randomizing which criteria are scored. We also measure what a policy does \emph{to} the rubric out of distribution.

\paragraph{Reward hacking and over-optimization.} That optimizing a proxy
degrades the true objective is one of the oldest observations in alignment
\citep{amodei2016concrete, skalse2022reward, pan2022misspecification}.
\citet{gao2023overoptimization} made this observation quantitative for learned reward models and showed that gold reward rises, peaks, and then decays as optimization proceeds. The mitigations developed in that literature operate on the reward model itself, by ensembling several such models \citep{coste2024ensembles, eisenstein2024herding}, averaging their weights \citep{rame2024warm}, or disentangling the hackable length component \citep{chen2024odin}. All of these train and serve extra reward models. Sampling sub-rubrics instead yields an implicit ensemble of
sub-objectives with no extra judge calls.
Out of distribution, we observe the same Gao-style rise and decay for rubric rewards. A concurrent work by \citet{mahmoud2026rewardhacking} documents rubric reward hacking as a divergence between the training verifier and a stronger judge. They attribute it to verifier failure and rubric limitations. CHERRL \citep{wang2026cherrl} reproduces the same failure in a controlled setting. It injects known biases into the judge and detects when the policy exploits them. Both studies diagnose rubric reward hacking.
We also evaluate rubric reward hacking out of distribution and propose a mitigation.

\paragraph{Regularization and reweighting.} Neuron dropout prevents co-adaptation by ensuring that no single unit can be relied on \citep{srivastava2014dropout}.
We apply the same idea to the training objective, so that the policy cannot rely on any single criterion. Dropout-style masking also appears elsewhere in the training loop. Vocabulary Dropout masks the proposer's output vocabulary to prevent diversity collapse in self-play \citep{dineen2026vocabdropout}, while we mask the reward's criteria.
Other work goes in the opposite direction. POW3R \citep{tyagi2026pow3r} \emph{reweights} criteria by the rollout group's verdict variance, so the most discriminative criteria receive the most optimization pressure. 
Unlike POW3R, dropout limits how often any single criterion can influence the updates. The
comparison in \cref{sec:pow3r} favors dropout.
GDPO \citep{liu2026gdpo} normalizes reward components separately within the group and targets signal resolution, not hacking. Dropout is orthogonal to both methods and only changes which criteria are scored.

%% file: sections/03_method.tex
\section{Method}
\label{sec:method}


\paragraph{Preliminaries.}
For a query $x$ and response $y$, a rubric consists of $K$ criteria
indexed by $k$, each with a weight $w_k$. A single judge call evaluates
all criteria and returns $s_k(x,y)\in\{0,1\}$, indicating whether
criterion $k$ is satisfied. Following RubricHub \citep{li2026rubrichub},
the reward is the normalized weighted score
\begin{equation}
R(x,y)
=
\frac{\sum_{k=1}^{K} w_k s_k(x,y)}
     {\sum_{k=1}^{K} w_k}.
\label{eq:scalar}
\end{equation}

\subsection{Measuring reward hacking}
\label{sec:protocol}

We define \emph{reward hacking} as a policy learning to cater to the training judge's preferences. Such a policy receives a high reward from that judge but shows a large proxy--gold score gap on out-of-distribution (OOD) test scenarios.
Our protocol uses two
judges and an OOD evaluation set. Every 20 training steps, we evaluate the current policy on the OOD evaluation set and grade each response twice, once with the training (\emph{proxy}) judge and once with a stronger, cross-family (\emph{gold}) judge.
We track four quantities: the \emph{gold score} (our best available estimate of true quality), the \emph{proxy$-$gold gap} (how much the proxy overrates the policy), the \emph{overclaim fraction} (the fraction of criteria the proxy marks satisfied but gold rejects), and the \emph{in-domain full-rubric reward} (what training optimizes). We use the in-domain full-rubric reward to check that a mitigation is not just slowing training. The gold judge is a stronger model, but it is not ground truth.
We therefore never interpret the size of the gap. A judge with a fixed bias shifts its curve by a constant.
It cannot, however, make the gold curve fall while the proxy curve rises. Our hacking signal is the divergence between the two curves during training, and a fixed bias in either judge can neither create nor remove it.

\subsection{Rubric Dropout}
\label{sec:dropout}

The method has a single hyperparameter, the dropout fraction $f\in[0,1)$. At
each training step we drop a random $f$-fraction of the rubric's criteria, always keeping at least three, and compute the same
reward on the kept criteria only. Because the number of dropped criteria is rounded and at least three are kept, the realized fraction is only approximately $f$, so we label the runs $f{\approx}30\%$ and $f{\approx}50\%$. Writing $m\in\{0,1\}^K$ for the keep-mask
($m_k=1$ means
criterion $k$ is kept),
\begin{equation}
  \tilde R(x,y;m) = \frac{\sum_{k} m_k\, w_k\, s_k(x,y)}{\sum_{k} m_k\, w_k}.
  \label{eq:drop}
\end{equation}
Evaluation always scores the full rubric with
\cref{eq:scalar}.
The judge grades all $K$ criteria in one call, so the full-rubric reward remains available for logging at no extra cost.

\subsection{GRPO with Rubric Dropout}
\label{sec:grpo}

We train with GRPO \citep{shao2024deepseekmath}.
For each prompt it samples a group of $G$ responses from the previous policy and computes a reward $R_i$ for each one. This reward is $R(x,y_i)$ for the base run and $\tilde R(x,y_i;m)$ under dropout. It then standardizes the rewards within the group into advantages $\hat A_i=(R_i-\mu)/\sigma$, where $\mu$ and $\sigma$ are the mean and standard deviation of the group's rewards, before the usual clipped policy-gradient update.

If each rollout drew its own mask, the $G$ responses would be graded on different sub-rubrics, and comparing them within the group would be meaningless. So we draw \emph{one mask per rollout group}, and the mask changes only between training steps. In practice, the mask's random seed is a hash of the prompt's instance id and the global training step. All rollouts of a prompt within a step therefore share one mask, and the mask is redrawn at every step.

The whole group shares one mask, so any reward normalizer that depends only on the mask cancels in GRPO's standardized advantage. The choice of normalizer in \cref{eq:drop} therefore does not affect the update (details in \cref{sec:analysis}). Over the mask distribution, dropout leaves the expected advantage unchanged up to a global scale that standardization removes.
Dropout's effect is instead the noise it adds to the advantage. This noise is largest for responses whose advantage hinges on a single criterion and small for responses that are broadly better than their group. This is the anti-co-adaptation argument behind neuron dropout, applied here to rubric criteria. We treat this as motivating
intuition rather than an established mechanism (\cref{sec:discussion}).




\subsection{Dropping correlated criteria together}
\label{sec:blockmethod}

The criterion-level dropout from \cref{sec:dropout} reduces overfitting by randomly dropping individual rubric criteria. In some datasets, however, correlations among criteria or annotation artifacts may cause multiple criteria to encode overlapping signals.
For these datasets, we also propose block dropout, which groups related criteria and drops them jointly.
Suppose a requirement is repeated across $r$ criteria, each dropped with probability $f$. Independent dropout removes the requirement entirely only with probability $f^r$, while dropping the $r$ criteria as one block removes it with probability $f$. Since $f^r < f$ for $r>1$, independent dropout is more likely to leave a
redundant copy active.\clnote{2026-09-25: display equation folded into the sentence, per Minglai.} We therefore group criteria expressing similar
requirements and apply one mask to each group.
Offline, we partition each rubric into blocks by grouping criteria whose text is semantically similar (\cref{sec:setup}; two worked examples in \cref{fig:block-example-sci,fig:block-example-med}). Let
$\{\mathcal B_b\}_{b=1}^{B}$ denote the resulting partition, and let
$M_b\in\{0,1\}$ indicate whether block $\mathcal B_b$ is retained.\clnote{2026-09-25: construction sentence moved ahead of the notation; embedding model, cosine threshold and size cap are in App.~B (Block construction).} Define
the retained criteria as
$\mathcal K_M=\bigcup_{b:M_b=1}\mathcal B_b$. The masked reward is
\begin{equation}
R_M(x,y)
=
\frac{\sum_{k\in\mathcal K_M} w_k s_k(x,y)}
     {\sum_{k\in\mathcal K_M} w_k}.
\label{eq:block-reward}
\end{equation}
The mask is shared by all responses in a rollout group, so the denominator
is constant within the group and cancels under GRPO normalization
(\cref{prop:norm}).
All criteria in a block are therefore added or removed as a unit, so redundant criteria cannot keep the same reward signal active.
During training, we visit blocks in random order and drop them until reaching the
target weight fraction. 

%% file: sections/04_experiments.tex
\section{Experiments}
\label{sec:experiments}

\subsection{Setup}
\label{sec:setup-brief}


\begin{wrapfigure}{r}{0.42\textwidth}  
  \vspace{-\baselineskip}
  \centering
  \includegraphics[width=0.40\textwidth]{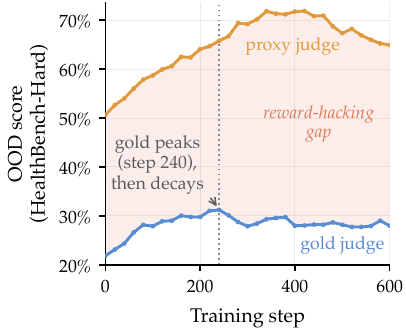}
  \caption{\textbf{The phenomenon.} Base Qwen3-8B on Medical, graded in-loop by both judges. The proxy climbs to 72\% while gold peaks and declines.}
  \label{fig:hacking}
  \vspace{-0.5\baselineskip}
\end{wrapfigure}
We train Qwen3 models ranging from 1.7B to 8B parameters \citep{qwen2025qwen3} with GRPO, using 16 rollouts per prompt and a learning rate of $10^{-6}$.
Our primary experiments use Qwen3-8B on two independent train$\to$eval pairs: RubricHub-Medical $\to$ HealthBench-Hard \citep{arora2025healthbench} and RubricHub-Science $\to$ ResearchQA \citep{researchqa2025}. HealthBench-Hard contains 1{,}000 prompts with physician-written rubrics. For ResearchQA, we report results on the 368 validation prompts that do not appear in the training data (\cref{sec:decontamination}). The proxy judge is \gpt{}, and the gold judge is \claude{}.
Throughout, we mainly compare \textbf{base} (no dropout), \textbf{30\%}, and \textbf{50\%} dropout. The ablations (\cref{sec:ablations}) include additional Medical runs with dropout fractions of 20--60\% and with POW3R.
Because hacking grows with training time, all cross-run results use a common 600-step horizon and a fixed comparison window over steps 400--600. We also compute matched-checkpoint win counts between runs on the same steps and prompts. Block dropout (\cref{sec:blockmethod}) uses $f{\approx}50\%$ with the same settings (\cref{sec:block}). Full details are provided in \cref{sec:setup}.

\subsection{Out of Distribution Reward Hacks}
\label{sec:phenomenon}
\Cref{fig:hacking} shows the base run on the Medical pair. In the first
phase, proxy and gold rise together, indicating the policy is genuinely
improving.
Around step 240, gold peaks at 31.2\% and then declines, while the proxy keeps rising to 72\%. The proxy$-$gold gap widens from 29\% to
44\%. By step 600, the policy has regressed below its step-240 level,
despite 360 more steps of ``improvement'' according to the proxy.
The Science pair shows the same divergence with a steeper collapse. Gold falls 22 points from its peak within 600 steps (the base curves in \cref{fig:science} and \cref{fig:reduce}).
\citet{gao2023overoptimization} established this over-optimization pattern for learned reward models. We see the same pattern with rubric rewards and in out-of-distribution settings, where the damage is largest.

\subsection{Dropout improves OOD scores by up to 7 points}\label{sec:main}

\begin{figure}[t]
  \centering
  \includegraphics[width=\linewidth]{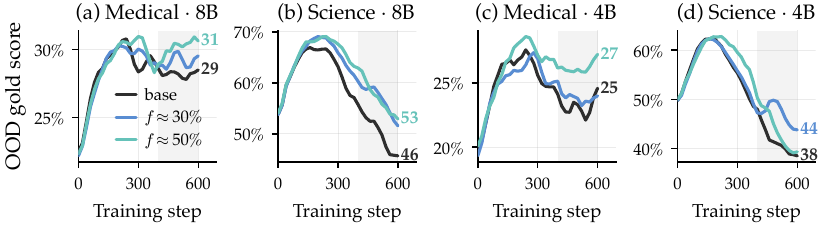}
  \caption{OOD gold score on both pairs at 8B and 4B (1.7B in \cref{fig:17b}).
  {\color{scaleInk}\rule[0.15ex]{1.2ex}{1.2ex}}~base,
  {\color{scaleBlue}\rule[0.15ex]{1.2ex}{1.2ex}}~$f{\approx}30\%$,
  {\color{scaleTeal}\rule[0.15ex]{1.2ex}{1.2ex}}~$f{\approx}50\%$.}
  \label{fig:science}
\end{figure}

\input{tables/table_main}

If these shortcuts stay stable because the rubric is fixed, resampling the rubric at every step should blunt the decline. \Cref{fig:science} and
\cref{tab:main} compare base with the two dropout runs on Medical and
Science over the same comparison window.

\textbf{Rubric dropout mitigates reward hacking across scales.}
We observe the same qualitative effect across models ranging from 1.7B to 8B parameters on both Medical and Science (\cref{fig:science,fig:17b,tab:main}). Rubric dropout consistently improves OOD performance late in training while preserving in-domain full-rubric reward. It also has little effect on the best OOD score attained during training. The trajectories begin to separate after their peaks, when the base runs undergo a larger decline.

\textbf{The improvement extends to both gold reward and hacking measures.}
For every model size and domain, at least one dropout setting achieves a higher late-window Gold score than base.
Peak scores remain comparable across runs and sometimes improve under dropout. The largest Gold difference between base and dropout occurs on Science at 8B. All three runs peak near step 200, but base then falls 21.5 points, to about 46\% at step 600. The declines under $f{\approx}30\%$ and $f{\approx}50\%$ are 18.0 and 16.4 points, respectively, and both runs finish near 53\%. Dropout also lowers overclaim in every model--domain setting and reduces the proxy--gold gap except on Medical at 1.7B (\cref{fig:reduce,tab:main}). These results indicate that the higher Gold scores are accompanied by weaker evidence of reward hacking.

\textbf{The preferred dropout fraction varies with model size and domain.}
At 8B, $f{\approx}50\%$ performs best on both Medical and Science.
At smaller scales the ordering is mixed. At 4B, $f{\approx}50\%$ performs better on Medical, whereas $f{\approx}30\%$ performs better on Science; at 1.7B, the preferred fraction again differs by domain. For the model sizes we tested, the higher dropout fraction appears more favorable at 8B. We treat this as a pattern in the present experiments only. The results do not establish a general relationship between model scale and the optimal dropout fraction.
If we view the same trajectories as a tradeoff between quality and hacking, the dropout runs match or exceed the base run on gold at matched overclaim levels (\cref{sec:setup}). 
\Cref{sec:discussion} discusses the mechanism behind the gain.

\subsection{Criterion-level breakdown}
\label{sec:where-gain}
\input{tables/table_breakdown}

To see where the aggregate changes come from, we break down the step-600 responses by criterion. Both judges grade every criterion for each run, and we classify each criterion accepted by the proxy as either confirmed by the gold judge or overclaimed (\cref{fig:rejudge} in the appendix). At matched proxy pass rates (within 1.3 points everywhere), both dropout runs have a higher gold pass rate and less overclaim. Gold pass rate increases and overclaim decreases monotonically with the dropout fraction. At $f{\approx}50\%$ the gains in gold pass rate are up to $+3.6$ points on Medical and $+7.3$ on Science. The proxy's error is also almost entirely one-sided. Underclaim is never above 3.1\%, so the gap comes from systematic over-crediting.
When we group criteria by their benchmark-provided tags (\cref{tab:axes}), the gains concentrate in criteria that ask for more substantive content. On Medical, the clinical axes (accuracy, completeness, context-awareness) gain the most, while the communication axes gain less. The largest single jump is on context-awareness at $f{\approx}50\%$, which is also the axis where the base run scores lowest. On Science the analytical types (comparison, limitation, impact) gain two to three times as much as the example and generic ones at $f{\approx}50\%$. The analytical-over-generic pattern also holds at $f{\approx}30\%$. In our runs, dropout preserves quality on the same expensive criteria that degrade first under hacking.
At 4B the same breakdown picks out the better fraction in each domain ($f{\approx}50\%$ on Medical, $f{\approx}30\%$ on Science). That fraction gains 3.9 (Medical) and 5.2 (Science) points of gold pass rate and lowers overclaim by 3.2 and 5.5. The other fraction stays near base. The matched checkpoint is step 600, the edge of the comparison window, so these numbers break down the results in \cref{tab:main} and do not re-estimate them.

\subsection{The separation persists for two epochs}
\label{sec:twoepoch}

\begin{figure}[t]
  \centering
  \includegraphics[width=\linewidth]{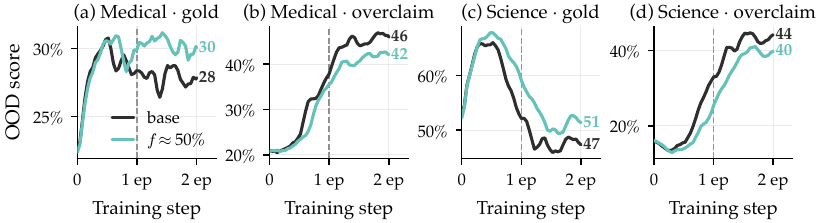}
  \caption{
\textbf{Base vs $f{\approx}50\%$ over two full epochs at 8B.}
Gold and overclaim on the Medical and Science pairs. The dashed line marks one epoch; curves are rolling means, with end values annotated.}
  \label{fig:2epoch}
\end{figure}

All results above use a 600-step training horizon, corresponding to roughly 1.3 epochs. This leaves open the possibility that dropout merely postpones collapse. We therefore extended the baseline and the \(f\approx50\%\) dropout run to two full epochs on both pairs. The gap persists throughout extended training (\cref{fig:2epoch}). After the peak, dropout consistently preserves more gold behavior and produces fewer overclaims on both Medical and Science. Because the two runs achieve similar peak scores, the main benefit of dropout lies in greater resistance to degradation during continued training.

\subsection{Block dropout narrows the leak}
\label{sec:block}

\begin{figure}[t]
  \centering
  \includegraphics[width=0.8\linewidth]{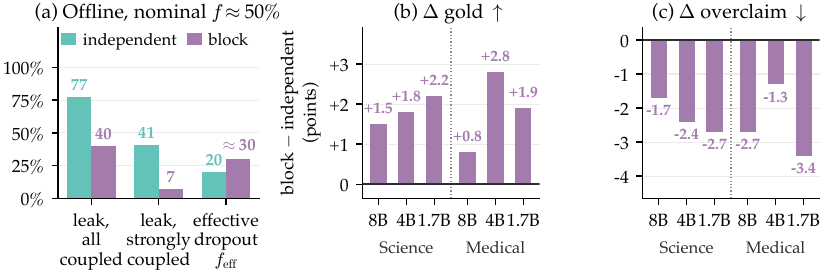}
  \caption{Block dropout: offline leak and training deltas. (a) Offline, on 8B Science verdicts at nominal $f{\approx}50\%$. (b, c) Block $-$ independent in training,
  window means over steps 400--600. }
  \label{fig:blockstory}
  \vspace{-1.3\baselineskip}
\end{figure}

We first measure the leak offline, replaying both kinds of mask
at nominal $f{\approx}50\%$ on logged 8B Science verdicts from a deep-plateau
checkpoint (\cref{tab:block}, left; \cref{fig:blockstory}a). Leakage
measures how much correlated reward signal survives a mask.
For each dropped criterion we take its largest verdict correlation $\phi$ with any retained criterion, then average over dropped criteria. Higher leakage means a near-copy of the dropped signal is still being rewarded. We use leakage only to evaluate a mask; the blocks themselves are built from text similarity alone (\cref{sec:blockmethod}).
Among criteria whose verdicts vary within the group, leakage falls from 41\% to 7\%. That row, however, uses oracle blocks. They are built from the verdicts themselves ($\phi\ge0.6$), not from the text blocks used in training, so the row only bounds what correlation-aware blocks could recover.\clnote{2026-09-25: operational definition of leakage added; stated that leakage is an evaluation metric, not the block-construction rule. Also disclosed that the 41$\to$7 row uses verdict-derived (oracle) blocks, as the appendix already says; the old sentence read as if text blocks achieved it.} The effective dropout rate is the share of
reward variance the mask removes; it rises from 20\% to about 30\%
(\cref{sec:setup}).


In training (\cref{tab:block,fig:blockstory}), block dropout consistently improves gold and reduces overclaim relative to independent dropout across both Science and Medical and at every model size. Effective dropout alone does not explain these gains. At matched effective dropout, performance depends on which criteria are dropped together. We discuss the reason in \cref{sec:discussion}.

%% file: tables/table_main.tex
\begin{table}[t]
  \centering
  \footnotesize
  \setlength{\tabcolsep}{1.7pt}
  \renewcommand{\arraystretch}{1.15}
  \resizebox{\linewidth}{!}{%
\begin{tabular}{clcccccc@{\hspace{9pt}}cccccc}
    \toprule
    & & \multicolumn{6}{c}{\textbf{Medical $\to$ HealthBench-Hard}}
    & \multicolumn{6}{c}{\textbf{Science $\to$ ResearchQA}} \\
    \cmidrule(lr){3-8}\cmidrule(l){9-14}
    & \textbf{Run} & Peak & Gold & $\Delta$ & Proxy$-$gold & Overclaim & Train reward
                 & Peak & Gold & $\Delta$ & Proxy$-$gold & Overclaim & Train reward \\
    \midrule
    & base
      & $31.2$ & $28.2$ & $\phantom{+}0.0$ & $40.3$ & $40.4$ & $\mathbf{98.0}$
      & $67.5$ & $50.4$ & $\phantom{+}0.0$ & $37.2$ & $37.3$ & $94.8$ \\
    \rowcolor{scaleBlue!13}
    & $f{\approx}30\%$
      & $30.9$ & $29.2$ & $+1.0$ & $38.7$ & $38.9$ & $97.8$
      & $69.4$ & $56.8$ & $+6.4$ & $29.9$ & $29.8$ & $\mathbf{96.3}$ \\
    \rowcolor{scaleTeal!16}
    \multirow{-3}{*}{\emph{8B}} & $f{\approx}50\%$
      & $\mathbf{31.5}$ & $\mathbf{30.1}$ & $\mathbf{+2.0}$ & $\mathbf{38.4}$ & $\mathbf{37.2}$ & $97.6$
      & $\mathbf{69.8}$ & $\mathbf{57.4}$ & $\mathbf{+7.0}$ & $\mathbf{29.5}$ & $\mathbf{29.5}$ & $95.7$ \\
    \midrule
    & base
      & $27.9$ & $23.2$ & $\phantom{+}0.0$ & $46.9$ & $47.1$ & $96.9$
      & $63.2$ & $41.6$ & $\phantom{+}0.0$ & $46.6$ & $46.7$ & $96.2$ \\
    \rowcolor{scaleBlue!13}
    & $f{\approx}30\%$
      & $28.0$ & $23.9$ & $+0.7$ & $\mathbf{43.2}$ & $45.9$ & $\mathbf{97.1}$
      & $62.6$ & $\mathbf{47.0}$ & $\mathbf{+5.3}$ & $\mathbf{40.1}$ & $\mathbf{40.3}$ & $96.7$ \\
    \rowcolor{scaleTeal!16}
    \multirow{-3}{*}{\emph{4B}} & $f{\approx}50\%$
      & $\mathbf{28.8}$ & $\mathbf{26.2}$ & $\mathbf{+3.0}$ & $44.4$ & $\mathbf{41.2}$ & $96.8$
      & $\mathbf{63.4}$ & $43.7$ & $+2.1$ & $43.2$ & $43.4$ & $\mathbf{96.8}$ \\
    \midrule
    & base
      & $19.8$ & $16.3$ & $\phantom{+}0.0$ & $\mathbf{50.0}$ & $57.6$ & $\mathbf{97.2}$
      & $50.3$ & $34.4$ & $\phantom{+}0.0$ & $48.1$ & $48.3$ & $\mathbf{92.7}$ \\
    \rowcolor{scaleBlue!13}
    & $f{\approx}30\%$
      & $\mathbf{20.1}$ & $\mathbf{18.0}$ & $\mathbf{+1.7}$ & $50.7$ & $\mathbf{53.2}$ & $97.0$
      & $\mathbf{53.6}$ & $36.8$ & $+2.4$ & $45.2$ & $45.4$ & $92.5$ \\
    \rowcolor{scaleTeal!16}
    \multirow{-3}{*}{\emph{1.7B}} & $f{\approx}50\%$
      & $19.8$ & $16.3$ & $\phantom{+}0.0$ & $52.5$ & $55.3$ & $96.6$
      & $52.3$ & $\mathbf{37.8}$ & $\mathbf{+3.4}$ & $\mathbf{43.9}$ & $\mathbf{44.1}$ & $92.2$ \\
    \bottomrule
  \end{tabular}}
  \caption{
Base and two dropout settings on both pairs at three model sizes. Gold (OOD) and Train reward (in-domain) are means over steps 400--600; Peak is the best gold score over all 600 steps. $\Delta$ is the point difference from each block's base. Bold marks column bests within each block.}
  \label{tab:main}
\end{table}

%% file: tables/table_breakdown.tex
\begin{table}[tb]
  \centering
  \footnotesize
  \setlength{\tabcolsep}{4.5pt}
  \renewcommand{\arraystretch}{1.15}
  \resizebox{\linewidth}{!}{%
\begin{tabular}{clccc@{\hspace{9pt}}cccccc}
    \toprule
    & & \multicolumn{3}{c}{\textbf{Medical, by HealthBench axis}}
    & \multicolumn{6}{c}{\textbf{Science, by ResearchQA rubric type}} \\
    \cmidrule(lr){3-5}\cmidrule(l){6-11}
    & \textbf{Gold pass rate}\,{\color{scaleGray}$\uparrow$} & clinical & context-aw. & communic.
                 & comparison & limitation & impact & other & example & citation \\
    \midrule
    & base
      & $50.2$ & $31.3$ & $35.4$
      & $41.9$ & $52.4$ & $48.9$ & $50.9$ & $59.5$ & $0.5$ \\
    \rowcolor{scaleBlue!13}
    & $f{\approx}30\%$
      & $53.6$ & $33.8$ & $\mathbf{37.6}$
      & $50.2$ & $61.9$ & $\mathbf{58.3}$ & $\mathbf{56.5}$ & $63.0$ & $0.0$ \\
    \rowcolor{scaleTeal!15}
    \multirow{-3}{*}{\emph{8B}} & $f{\approx}50\%$
      & $\mathbf{54.1}$ & $\mathbf{44.0}$ & $35.9$
      & $\mathbf{55.9}$ & $\mathbf{64.8}$ & $58.0$ & $55.5$ & $\mathbf{63.4}$ & $0.5$ \\
    \midrule
    & base
      & $44.5$ & $34.0$ & $33.7$
      & $35.5$ & $\mathbf{51.0}$ & $41.2$ & $41.1$ & $47.9$ & $0.3$ \\
    \rowcolor{scaleBlue!13}
    & $f{\approx}30\%$
      & $44.4$ & $26.7$ & $\mathbf{37.0}$
      & $\mathbf{41.9}$ & $47.4$ & $\mathbf{47.9}$ & $\mathbf{45.9}$ & $\mathbf{54.0}$ & $\mathbf{0.8}$ \\
    \rowcolor{scaleTeal!15}
    \multirow{-3}{*}{\emph{4B}} & $f{\approx}50\%$
      & $\mathbf{48.5}$ & $\mathbf{34.5}$ & $36.9$
      & $34.5$ & $50.5$ & $43.7$ & $40.6$ & $48.0$ & $0.3$ \\
    \bottomrule
  \end{tabular}}
  \caption{\textbf{Gold pass rate by HealthBench axis and ResearchQA rubric
  type} (\%, at step 600).
  Context-awareness is a
  subset of the clinical axes.
Bold marks the best run per column within each block.
  Criteria with multiple type tags are counted under each of their types.}
  \label{tab:axes}
\end{table}

%% file: sections/05_ablations.tex
\section{Ablations}
\label{sec:ablations}

\subsection{The dropout fraction}
\label{sec:dose}

\begin{wrapfigure}{r}{\dosewrapwidth\textwidth}  
  \vspace{-5\baselineskip}
  \centering
  \includegraphics[width=\dosefigwidth\textwidth]{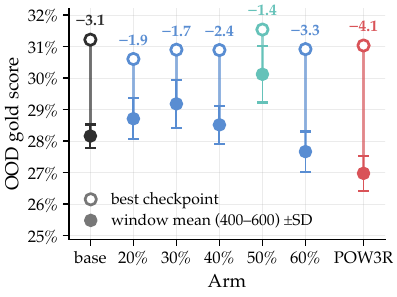}
  \caption{\textbf{Sweeping the dropout fraction (Medical).} The post-peak drop is
  smallest at $f{\approx}50\%$.}
  \label{fig:dose}
  \vspace{\dosewrapbottom\baselineskip}  
\end{wrapfigure}

\looseness=-1
We sweep $f\in\{20,30,40,50,60\}\%$
(\cref{fig:dose}; full numbers in \cref{tab:ablation} in the appendix). Best-checkpoint gold is essentially
tied across all runs (30.6--31.5\% against base's 31.2\%), so the
differences come from how gold holds up under continued training. Everything from 20\% to 50\% is at or above base. The best window mean is at 50\% ($+2.0$).
The sign flips only at 60\% ($-0.5$). We expected this failure once the surviving sub-rubrics no longer cover all aspects of quality. The training trajectories (\cref{fig:traj}, appendix) confirm that these differences come from the post-peak phase and not from learning speed. They also show that no run gives up in-domain full-rubric reward. Any dropout fraction in the 30--50\% range gives most of the benefit.

\subsection{Reweighting is not a substitute for dropping}
\label{sec:pow3r}

The natural alternative to dropping criteria, reweighting toward the informative
ones, performs worse out of distribution than no intervention at all.
POW3R has the lowest OOD gold score of any run (27.0\%) and loses to base at all matched checkpoints. Its overclaim fraction, 42.2\%, is also the highest, above even base's 40.4\%. One plausible mechanism is that reweighting by verdict variance concentrates pressure on exactly the criteria the policy is learning to game. This would then strengthen the loop that dropout weakens. POW3R's best checkpoint (31.0\%) matches base's (31.2\%), so the deficit comes from how its score decays after the peak. We test it OOD at its published defaults and do not evaluate
its in-distribution claims.
Both the sweep and the POW3R comparison indicate that OOD robustness improves when optimization pressure is spread across criteria and degrades when it is concentrated.

%% file: sections/06_discussion.tex
\section{Discussion}
\label{sec:discussion}

\newcommand{\entropywrap}{%
\begin{wrapfigure}{r}{0.4\textwidth}
  \vspace{-\baselineskip}
  \centering
  \includegraphics[width=0.4\textwidth]{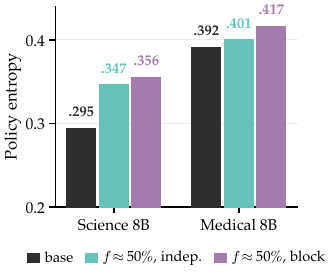}
  \caption{\textbf{Policy entropy at 8B}, mean over training steps
  300--600. Both dropout variants stay above base on both pairs.}
  \label{fig:entropy}
  \vspace{-0.5\baselineskip}
\end{wrapfigure}}

\textbf{Where the gain comes from.} \cref{sec:where-gain} shows that the
gain comes from the prompt-specific, content-heavy criteria, which dropout
keeps satisfied after the base policy stops satisfying them.
We attribute the gain on these criteria to the same mechanism that makes neuron dropout work. In that setting, dropout prevents units from co-adapting, that is, from relying on specific partners that are always present \citep{srivastava2014dropout}. With the rubric resampled at every
step, no criterion is reliably present. A shortcut that satisfies a generic
criterion, such as a fixed template, is rewarded only on the steps where
that criterion is kept, so it is never reliably reinforced.
To score reliably, the policy has to answer the prompt, which is what the prompt-specific criteria measure.\clnote{2026-09-25: explains anti-co-adaptation and adds the missing link from unreliable shortcuts to the prompt-specific criteria.}

\textbf{A better tradeoff, not the same path.}
We compare the two-epoch runs of \cref{sec:twoepoch} at matched overclaim. The comparison is informative only after one epoch, once each prompt has been scored under different masks. Past that point the $f{\approx}50\%$ run holds more gold than base at every matched point on Medical ($+0.2$ to $+2.5$ points) and at nearly every matched point on Science ($+1.6$ on average). Dropout reaches a better gold-versus-overclaim tradeoff rather than the same one later. With one run per condition, we read this as supporting evidence.

\entropywrap
\textbf{Mask structure matters.} At the same nominal dropout fraction, block dropout outperforms independent dropout on both domains at every model size (\cref{sec:block}).
The difference comes from redundancy between criteria. When two criteria encode the same requirement, independent dropout usually removes only one of them, and the survivor keeps rewarding the same behavior. Dropping them together removes that signal, so the policy cannot keep exploiting it through a near-duplicate. Block masks also remove a larger share of reward variance (\cref{tab:block}).
The gains therefore depend on which criteria are dropped together as well as on how many.

\textbf{Output diversity.}
Late in training, base responses consist largely of template scaffolding (repeated headers and placeholder lists) that satisfies structural criteria but adds little content. On Science, every late-stage base response reaches the generation length limit. The dropout policies stay more diverse than the base policy. Single-step entropy is noisy, so we average it over training steps 300--600. Gold peaks near step 240, and base collapses during this later window. Over that span, policy entropy on Science is 0.295 for base, 0.347 for independent dropout and 0.356 for block dropout. Medical shows the same ordering with a smaller gap (\cref{fig:entropy}).\clnote{2026-09-25: the values are the mean over steps 300--600, not the final checkpoint (checked on W\&B). Science keeps base lowest in every window; on Medical the ordering depends on the window (over 400--600 independent dropout is below base, 0.359 vs 0.391).}
The base policy's collapse onto templates and its low entropy indicate overfitting to the fixed rubric. The dropout runs do not show this overfitting pattern. Their outputs stay varied and do not converge on a single template that satisfies the proxy.

%% file: sections/07_conclusion.tex
\section{Conclusion}

Rubric-based RL can overfit a fixed reward proxy. Across both domains, OOD gold scores peak and then decline even as proxy scores continue to rise.
Rubric Dropout counters this by randomly omitting criteria during training while keeping the full rubric for evaluation. Across models from 1.7B to 8B, it improves OOD gold scores and reduces overclaim without lowering in-domain reward. Dropout rates between 30\% and 50\% work well across most settings.
By contrast, variance-based reweighting performs worse than the no-dropout baseline.
Block dropout provides further gains by dropping semantically correlated criteria together. These results suggest that changing which criteria influence each update makes stable rubric shortcuts harder to exploit while promoting greater rollout diversity. Limitations are discussed in \cref{sec:limitations}.

%% file: sections/08_limitations.tex
\subsection*{Ethics Statement}
This work uses medical prompts and benchmarks as a testbed for reward hacking.
We do not release a model intended for clinical use. The failure mode we document, a policy that satisfies its training judge while true quality degrades, is itself a deployment risk for rubric-trained models. Measuring it is one of the aims of this work. All judges are commercial APIs used under their
terms. No human-subjects data were collected.

\subsection*{Reproducibility Statement}
All numbers derive from per-step trajectories logged during training and
from the judges' grades of saved model outputs. The window means, win counts, and figures are
computed from the cached data. Models, data,
judges, hyperparameters, and the dropout procedure, including block
construction, are specified in
\cref{sec:method} and \cref{sec:setup}.

%% file: sections/09_appendix.tex
\newpage
\renewcommand{\textfraction}{0.25}\renewcommand{\topfraction}{0.75}\renewcommand{\bottomfraction}{0.5}
\setcounter{topnumber}{2}\setcounter{bottomnumber}{1}\setcounter{totalnumber}{3}
\raggedbottom

\section{Analysis of Group-Shared Rubric Dropout}
\label{sec:analysis}

Group-shared Rubric Dropout is well-behaved under GRPO for two reasons. Fix
one group and one mask $m$, and write $s_{k,i}$ for the verdict of criterion $k$ on response
$y_i$ and $c_i=\sum_k m_k w_k s_{k,i}$ for the masked score of $y_i$ before
any normalization. Here $\mathrm{mean}_j$ and $\mathrm{std}_j$ run over the
group's responses $j=1,\dots,G$.

\needspace{10\baselineskip}
\begin{proposition}[The normalizer cancels]
\label{prop:norm}
Because the mask is shared by the whole group, any positive normalizer $Z$ that depends only on the mask (the kept weight $\sum_k m_k w_k$ of \cref{eq:drop}, its expectation, or no normalizer at all) is the same constant for every response in the group. Whenever the group's masked scores are not all equal, the normalizer therefore cancels in the standardized advantage:
\begin{equation*}
  \hat A_i(m)
  =\frac{c_i/Z-\mathrm{mean}_j(c_j/Z)}{\mathrm{std}_j(c_j/Z)}
  =\frac{c_i-\mathrm{mean}_j(c_j)}{\mathrm{std}_j(c_j)}.
\end{equation*}
Under this standardization, the advantage depends only on which criteria the mask keeps. In particular, there is no normalizer to tune.
\end{proposition}

The second reason is the intuition behind the method.
In expectation, dropout only rescales the advantage, so any effect it has comes from the noise it injects. This noise is largest for responses whose advantage depends on a single criterion. We take the mask entries to be i.i.d., keeping each criterion with probability $1-f$. Center each verdict within the group,
$\delta_{k,i}=s_{k,i}-\bar s_k$ with $\bar s_k=\frac1G\sum_{j=1}^G s_{k,j}$,
and consider the un-normalized advantage $u_i(m)=\sum_k m_k w_k \delta_{k,i}$.
With the full rubric this is the centered reward up to the constant total weight, $u_i(\mathbf{1}) = \big(\sum_k w_k\big)(R_i-\mu)$, and group standardization ignores that constant (\cref{prop:norm}).

\begin{observation}[Dropout is a variance regularizer]
\label{obs:var}
Over the i.i.d.\ mask distribution,
\begin{equation*}
  \mathbb{E}_m[u_i(m)]=(1-f)\,u_i(\mathbf{1}),\qquad
  \mathrm{Var}_m[u_i(m)]=f(1-f)\sum_k w_k^2\,\delta_{k,i}^2.
\end{equation*}
\end{observation}

In expectation, the only effect of dropout on the un-normalized advantage is a global scale $1-f$, which group standardization removes. Dropout therefore acts only through the variance term, and this term shows which responses get a noisy signal. It is largest exactly when the advantage hinges on one high-weight criterion (one large $w_k\delta_{k,i}$), and smallest when a response is broadly better than its group (many small contributions).
Since gradient noise slows SGD's progress along a direction, dropout slows progress toward single-criterion exploits more than toward broad improvement. This is the same anti-co-adaptation logic as neuron dropout, applied to the reward. We treat it as motivating intuition only; it is not an established mechanism. The experiments
establish where dropout helps, and \cref{sec:discussion} discusses what they
can and cannot say about why. There are two caveats to
\cref{obs:var}.
In practice we drop a fixed $f$-fraction rather than taking i.i.d.\ draws. Each criterion is then kept with probability $1-f$, so the expectation is unchanged, and the variance only acquires small cross-terms from the negative mask covariances. And because GRPO's division by the group standard
deviation also depends on the mask, \cref{obs:var} describes the advantage
before that division.

\Cref{obs:var} also shows at which dropout fraction the injected variance is largest. The factor
$f(1-f)$ peaks at $f=1/2$, so under the variance-regularization reading the
effect is strongest near 50\% dropout.
The Medical sweep agrees. Its best window mean is at $f{\approx}50\%$, and the benefit collapses at 60\%, where the kept sub-rubric no longer covers the criteria that define quality in the full rubric (\cref{sec:dose}). We note the agreement but do not rely on it.
We treat the variance-regularization reading as motivation only, since the sweep cannot separate it from the coverage effect that dominates at large $f$.

\subsection{Dropping correlated criteria together}
\label{sec:blockanalysis}\clnote{2026-09-24: inserted verbatim from Minglai. This copy carries a new label, since Sec.~3.4 already uses the original one.}

Independent criterion dropout may leave redundant reward signals active.
Suppose that $r$ criteria produce identical verdicts and
are independently dropped with probability $f$. Their common signal is
removed only if all $r$ criteria are dropped. If they belong to the same
block, dropping that block removes the signal at once:
\begin{equation}
\Pr(\text{common signal is removed})
=
\begin{cases}
f^r, & \text{independent criterion dropout},\\
f,   & \text{block dropout},
\end{cases}
\label{eq:block-survival}
\end{equation}
where the second line assumes that the $r$ criteria share a block. This
calculation uses Bernoulli dropout for illustration. With exactly $m$
of $K$ criteria dropped uniformly without replacement, the first
probability is
$\binom{K-r}{m-r}/\binom{K}{m}$.

We next characterize how block dropout affects within-group reward variation. For a
rollout group $\{y_i\}_{i=1}^{G}$, write
$s_{ik}=s_k(x,y_i)$ and define
\begin{equation}
z_{ik}
=
s_{ik}-\frac{1}{G}\sum_{j=1}^{G}s_{jk},
\qquad
T_{ib}
=
\sum_{k\in\mathcal B_b}w_k z_{ik},
\end{equation}
where $\{\mathcal B_b\}_{b=1}^{B}$ is a partition of the rubric. Given a
block mask $M=(M_1,\ldots,M_B)$, the centered reward numerator is
\begin{equation}
U_i(M)=\sum_{b=1}^{B}M_bT_{ib}.
\end{equation}
The retained-weight normalizer is shared by every response in the
rollout group and therefore cancels under GRPO normalization
(\cref{prop:norm}).

Consider the Bernoulli approximation in which each block is retained
independently with probability $q=1-f$. Define
$\operatorname{Var}_i$ and $\operatorname{Cov}_i$ using the population
normalization $1/G$. The mask-induced variance, averaged over responses,
is
\begin{align}
V_{\mathrm{block}}
&\coloneqq
\frac{1}{G}\sum_{i=1}^{G}
\operatorname{Var}_{M}\!\left[U_i(M)\right]
\notag\\
&=
V_{\mathrm{ind}}
+
2q(1-q)
\sum_{b=1}^{B}
\sum_{\substack{k<\ell\\k,\ell\in\mathcal B_b}}
w_k w_\ell
\operatorname{Cov}_i(s_{ik},s_{i\ell}),
\label{eq:block-var}
\end{align}
where
\begin{equation}
V_{\mathrm{ind}}
=
q(1-q)\sum_{k=1}^{K}
w_k^2\operatorname{Var}_i(s_{ik})
\end{equation}
is the corresponding quantity for independent criterion dropout. Hence
$V_{\mathrm{block}}\geq V_{\mathrm{ind}}$ when the within-block verdict
covariances are nonnegative.
Block dropout therefore perturbs correlated reward signals as a group, whereas under independent criterion dropout part of such a group often stays active. Negative within-block covariance can reverse the variance inequality, which motivates using a conservative similarity threshold and splitting clusters that exceed the size cap.

We construct the blocks once, offline, using criterion-text similarity.
At each training step, blocks are sampled without replacement and
dropped sequentially toward a target fraction $f$ of the total rubric
weight, subject to retaining at least three criteria.
Because blocks are indivisible, the realized retained-weight fraction can deviate slightly from its target; for $f\approx50\%$, it is around $50\%$ on average. Every response in a
rollout group shares the same mask.
Block dropout adds one offline clustering pass and requires no extra judge calls. \Cref{sec:block} reports the results.

\needspace{8\baselineskip}
\section{Experimental Setup and Details}
\label{sec:setup}

\paragraph{Policy and algorithm.}
We train Qwen3-8B \citep{qwen2025qwen3} with GRPO \citep{shao2024deepseekmath}, using 16 rollouts per prompt, a learning rate of $10^{-6}$, and FSDP. Qwen3-4B and Qwen3-1.7B are trained with the same recipe. All 8B runs train to at least 600 steps except the $f{\approx}60\%$ run, which ends at step 559 with its last logged evaluation at step 540. The smaller sizes are described under Training runs. All comparisons use the common 600-step horizon.

\paragraph{Training data.} RubricHub-Medical consists of medical prompts with
weighted rubrics of 8--67 criteria (mean $\sim$30). RubricHub-Science is built the same way and has 29{,}418 prompts with a mean of $\sim$27 criteria. The proxy judge for both is \gpt{}.

\paragraph{OOD evaluation.}
\label{sec:decontamination}
The Medical pair is evaluated on the Hard split of HealthBench \citep{arora2025healthbench}, which has 1{,}000 prompts with a mean of $\sim$11.9 criteria each. This split shares no prompts with RubricHub-Medical. The Science pair is evaluated on the ResearchQA validation split \citep{researchqa2025}. Since RubricHub-Science is built from ResearchQA, we score on the 368 prompts that never occur in training, with a mean of $\sim$7.4 equal-weight criteria per prompt. The
evaluation rubrics share no criteria with the training rubrics on any
prompt. Both pairs are evaluated in-loop every 20 steps and graded by the proxy (\gpt{}) and gold (\claude{}) judges as described in \cref{sec:protocol}. The in-loop Science evaluation runs on the full
703-prompt split, and the reported Science numbers come from scoring the saved
responses on the 368-prompt subset.
The full split shows the same trends, slightly weaker.

\begin{figure}[tb]
  \centering
  \includegraphics[width=0.97\linewidth]{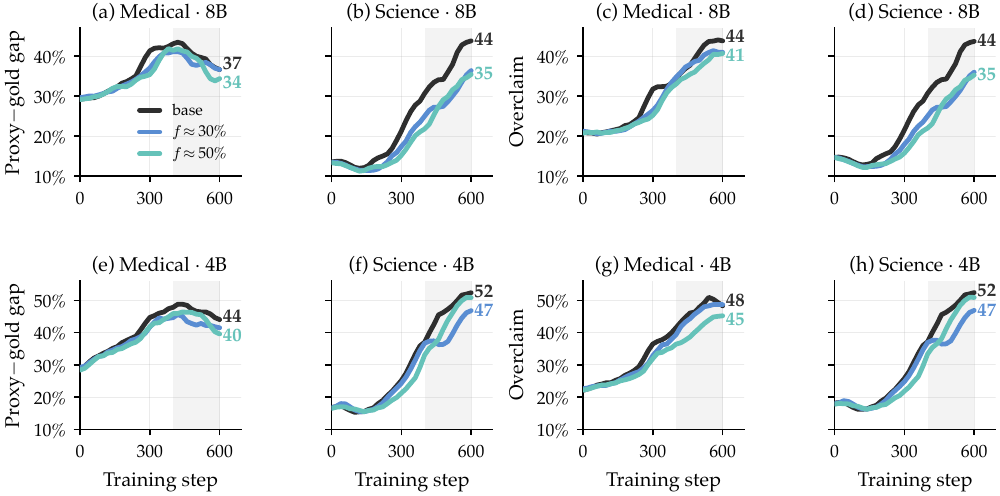}
  \caption{
\textbf{Dropout lowers both hacking measures in the Medical and Science domains at 8B and 4B} (for 1.7B, see \cref{fig:17b}). Proxy$-$gold gap and overclaim fraction.
  {\color{scaleInk}\rule[0.15ex]{1.2ex}{1.2ex}}~base,
  {\color{scaleBlue}\rule[0.15ex]{1.2ex}{1.2ex}}~$f{\approx}30\%$,
  {\color{scaleTeal}\rule[0.15ex]{1.2ex}{1.2ex}}~$f{\approx}50\%$.}
  \label{fig:reduce}
\end{figure}

\begin{figure}[tb]
  \centering
  \includegraphics[width=0.95\linewidth]{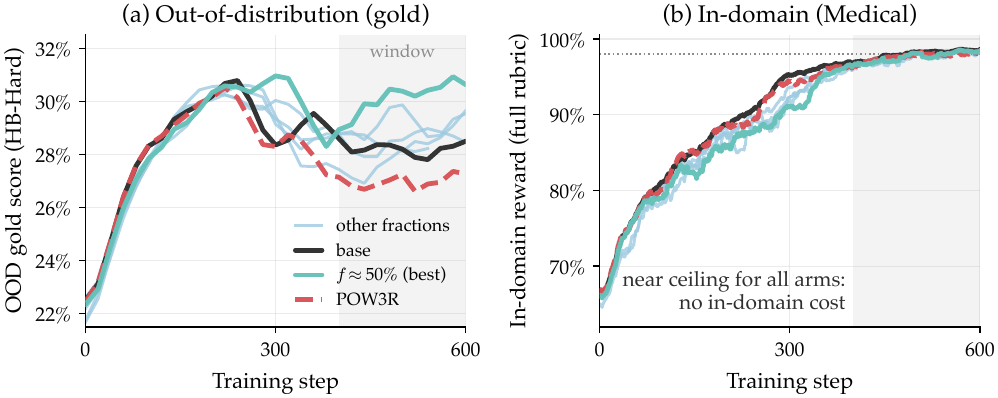}
  \caption{\textbf{Trajectory view of the sweep (Medical).} (a) OOD gold
  score, 3-point rolling means of the 20-step evaluations. (b) In-domain full-rubric
  reward, 25-step rolling mean of the per-step training signal. Window shaded.}
  \label{fig:traj}
\end{figure}

\input{tables/table_ablation}

\begin{figure}[tb]
  \centering
  \includegraphics[width=0.9\linewidth]{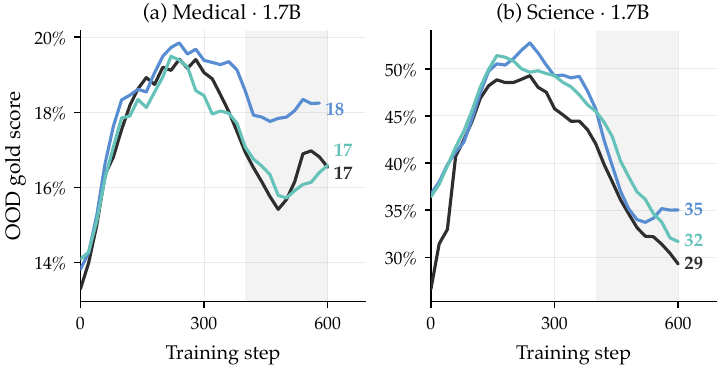}
  \caption{\textbf{Qwen3-1.7B trajectories on both pairs.}
  {\color{scaleInk}\rule[0.15ex]{1.2ex}{1.2ex}}~base,
  {\color{scaleBlue}\rule[0.15ex]{1.2ex}{1.2ex}}~$f{\approx}30\%$,
  {\color{scaleTeal}\rule[0.15ex]{1.2ex}{1.2ex}}~$f{\approx}50\%$.
  Rolling means, comparison window shaded.}
  \label{fig:17b}
\end{figure}

\begin{figure}[tb]
  \centering
  \includegraphics[width=0.97\linewidth]{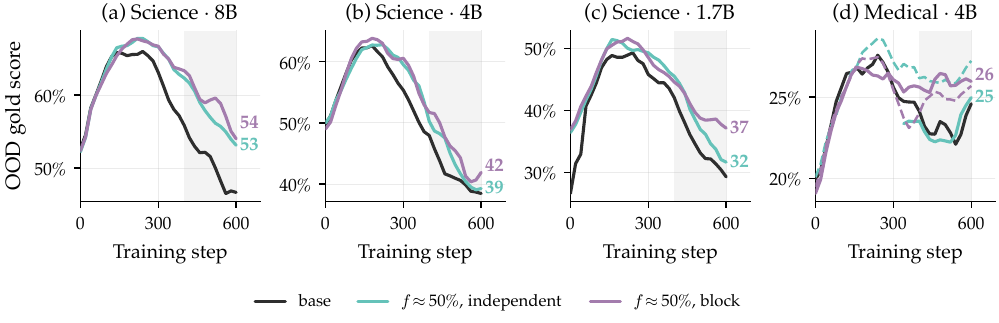}
  \caption{
\textbf{Block dropout against independent dropout} (Medical 4B has two runs of each dropout kind; solid lines are the runs compared in \cref{tab:block}, dashed lines the others).
  {\color{scaleInk}\rule[0.15ex]{1.2ex}{1.2ex}}~base,
  {\color{scaleTeal}\rule[0.15ex]{1.2ex}{1.2ex}}~$f{\approx}50\%$ independent,
  {\color{scalePurple}\rule[0.15ex]{1.2ex}{1.2ex}}~$f{\approx}50\%$ block.
  Rolling means, comparison window shaded.}
  \label{fig:block}
\end{figure}

\paragraph{Training runs.} On Medical we run base, $f\in\{20,30,40,50,60\}\%$,
and POW3R. On Science we run base and $f\in\{30,50\}\%$. For Qwen3-4B and
Qwen3-1.7B we run base and $f\in\{30,50\}\%$ on both pairs, with the same
recipe.
For 4B and 1.7B we report in-loop trajectory statistics on the same 368-prompt subset. Preemptions fragmented the 1.7B logs, and we
merge the segments by step. The 1.7B Medical $f{\approx}30\%$ run stops at step
599 with its last eval at step 580, so its window statistics cover steps
400--580; the other 1.7B Medical runs cover the full window. \Cref{fig:17b} shows the 1.7B trajectories.
The two-epoch continuations of \cref{sec:twoepoch} resume each 8B run from its deepest checkpoint. At the resume boundary, the continuation's gold score agrees with the original run's gold score to within 0.7 points, and the merged curves have no eval gaps. The block-dropout runs (\cref{sec:blockmethod,sec:block}) use $f{\approx}50\%$ and the recipe above on both pairs at 8B, 4B, and 1.7B. We score them in-loop like the other small-model runs. There is one run per cell except Medical 4B, where both dropout kinds have two runs. In \cref{tab:block} and \cref{fig:blockstory} we compare the later run of each dropout kind (window gold 25.9 for block vs 23.1 for independent). All four runs are drawn in \cref{fig:block}, with the earlier run of each kind dashed (block 25.1, independent 26.2). \Cref{fig:block} shows four of the six domain-by-size cells.\clnote{2026-09-23: Medical 4B block updated to the later run (gold 25.9, overclaim 44.1 vs independent 23.1, 45.4, Minglai's numbers): $+2.8$ gold, $-1.3$ overclaim, so block now wins on both there; the ``constant criteria'' explanation is dropped.}\clnote{2026-09-20: Medical 8B block added to \cref{tab:block} and \cref{fig:blockstory} from Minglai's number; its trajectory is not in \cref{fig:block} yet. Tell me which W\&B run(s) the $+0.8$/$-2.7$ come from (Medical 8B block runs exist for seeds 42, 43, 45, 46) and I will add the panel.}

\input{tables/table_block}

\paragraph{Block construction.} We embed each criterion with
\texttt{all-mpnet-base-v2} and compute cosine similarity between normalized
embeddings. We connect pairs with similarity at least $0.62$ and take
connected components as blocks.
The threshold $0.62$ is a conservative heuristic. We keep it fixed and do not tune it on downstream results. In a preliminary analysis at $0.60$, components often grouped criteria that shared a broad topic but were not redundant requirements. Because connected components
can also grow through transitive links, we split components larger than
$\max(3,\operatorname{round}(0.4K))$ into singletons.
The fixed size cap also limits over-merging of topically similar criteria. As with the threshold, we did not select the cap using downstream training results. At training time the group-shared mask drops whole blocks until the dropped fraction reaches $f$. It skips any block whose removal would leave fewer than three kept criteria. The last dropped block can overshoot the target, so the realized kept weight is around 50\% rather than exactly the nominal value. Evaluation is
unchanged.

\paragraph{Two block examples.}\label{app:block-examples}
\Cref{fig:block-example-sci,fig:block-example-med} show the text-similarity blocks for one rubric from each pair, produced by the procedure above and used as is in training. Criteria are shown verbatim; the italic label is our gloss of the requirement a block shares. Within a block the criteria restate one requirement in different words, so dropping them independently would usually leave a near-copy active. No verdicts or scores enter the grouping. Because whole blocks are dropped, a single mask can remove several criteria at once; the rule that every rubric keeps at least three criteria therefore binds more often on such rubrics than under independent dropout, and the kept weight lands near, rather than exactly at, the nominal 50\%.

\newcommand{\blk}[2]{\item[\textbf{#1}] \textit{#2}}
\newcommand{\crit}[1]{\item #1}
\begin{figure*}[!t]
\small
\begin{tcolorbox}[colback=scaleTeal!6!white,colframe=scaleTeal!70!black,boxrule=0.6pt,arc=2pt,left=6pt,right=6pt,top=4pt,bottom=4pt,
  title={\textbf{Example A: Science (RubricHub-Science).} Prompt: ``When a standard die (with numbers 1 to 6) is rolled, what is the probability that the number rolled is neither a multiple of 5 nor an odd number?'' (16 criteria: five blocks, four singletons)},fonttitle=\small\bfseries,coltitle=white]
\begin{description}[nosep,leftmargin=1.2em,font=\normalfont]
  \blk{Block 1}{states the sample space and uses correct terminology}
  \begin{itemize}[nosep,leftmargin=1.2em,label=--]
    \crit{Explicitly lists the six possible outcomes of a standard die: \{1, 2, 3, 4, 5, 6\}.}
    \crit{Uses correct and consistent terms such as `standard die', `outcome', `odd', `multiple of 5', and `probability' throughout.}
  \end{itemize}
  \blk{Block 2}{identifies the multiple-of-5 and odd outcomes}
  \begin{itemize}[nosep,leftmargin=1.2em,label=--]
    \crit{Correctly identifies 5 as the only multiple of 5 within the sample space.}
    \crit{Correctly identifies the odd outcomes in the sample space as \{1, 3, 5\}.}
  \end{itemize}
  \blk{Block 3}{interprets ``neither \ldots nor'' and builds the exclusion set}
  \begin{itemize}[nosep,leftmargin=1.2em,label=--]
    \crit{Interprets the phrase `neither a multiple of 5 nor an odd number' to exclude all numbers that are multiples of 5 and all odd numbers.}
    \crit{Identifies all numbers to be excluded: the odd set \{1, 3, 5\} and the multiple-of-5 set \{5\}.}
    \crit{Explicitly states that the element appearing in both the `multiple of 5' set and the `odd' set is counted only once, and provides a brief justification for this single counting.}
  \end{itemize}
  \blk{Block 4}{applies the probability formula and gives 1/2}
  \begin{itemize}[nosep,leftmargin=1.2em,label=--]
    \crit{Computes probability as (favorable outcomes)/(total outcomes) and simplifies the fraction correctly.}
    \crit{Uses the probability formula P = favorable/total to calculate the numeric value.}
    \crit{Provides the final probability as the simplified fraction 1/2 (or equivalent 0.5 or 50\%).}
  \end{itemize}
  \blk{Block 5}{organizes the solution step by step}
  \begin{itemize}[nosep,leftmargin=1.2em,label=--]
    \crit{Presents the solution in a logical, step-by-step order: state sample space, apply exclusion criteria, determine favorable outcomes, compute probability.}
    \crit{Organizes the solution into at least four clearly numbered sections with descriptive headings that follow the logical order: (1) Sample Space, (2) Excluded Numbers, (3) Favorable Outcomes, (4) Probability Calculation.}
  \end{itemize}
  \blk{Singletons}{}
  \begin{itemize}[nosep,leftmargin=1.2em,label=--]
    \crit{Lists exactly the outcomes that satisfy the condition, namely \{2, 4, 6\}, and does not include any excluded numbers.}
    \crit{Counts three favorable outcomes and six total outcomes without error.}
    \crit{Explanation is concise, free of irrelevant material, and easy to follow.}
    \crit{Displays the probability calculation inside LaTeX math delimiters (e.g., \texttt{\$\$P = \textbackslash frac\{3\}\{6\} = \textbackslash frac\{1\}\{2\}\$\$}), showing both the unsimplified and simplified fractions within the same LaTeX block.}
  \end{itemize}
\end{description}
\end{tcolorbox}
\caption{\textbf{Text-similarity blocks for a Science rubric.} Criteria are grouped offline by embedding similarity (\texttt{all-mpnet-base-v2}, cosine $\ge 0.62$, connected components); the italic label is our gloss of the requirement the block shares. Block dropout drops or keeps each block as a unit.}
\label{fig:block-example-sci}
\end{figure*}

\begin{figure*}[!t]
\small
\begin{tcolorbox}[colback=scaleBlue!6!white,colframe=scaleBlue!70!black,boxrule=0.6pt,arc=2pt,left=6pt,right=6pt,top=4pt,bottom=4pt,
  title={\textbf{Example B: Medical (RubricHub-Medical).} Prompt: ``What is the typical age range for children who commonly experience nursemaid's elbow or annular ligament displacement?'' (16 criteria: five blocks)},fonttitle=\small\bfseries,coltitle=white]
\begin{description}[nosep,leftmargin=1.2em,font=\normalfont]
  \blk{Block 1}{names the condition and its core age range (1--4 years, peak 2--3)}
  \begin{itemize}[nosep,leftmargin=1.2em,label=--]
    \crit{The response explicitly states that nursemaid's elbow most commonly occurs in children between 1 and 4 years of age.}
    \crit{The response specifies that the peak incidence of nursemaid's elbow is between 2 and 3 years of age.}
    \crit{The response correctly identifies nursemaid's elbow as radial head subluxation or annular ligament displacement (or uses an equivalent term).}
    \crit{Each of the two synonymous terms, ``radial head subluxation'' and ``annular ligament displacement'', appears at least twice in the response.}
  \end{itemize}
  \blk{Block 2}{the broader at-risk age span (about 6 months to 5--6 years)}
  \begin{itemize}[nosep,leftmargin=1.2em,label=--]
    \crit{The response mentions a broader at-risk age span, e.g., from about 6 months up to approximately 5--6 years.}
    \crit{The response states the peak incidence as the exact numeric interval ``2 to 3 years'' (inclusive) in a separate sentence.}
    \crit{The response provides the broader at-risk span with both lower and upper numeric bounds (e.g., ``6 months'' and ``5--6 years'') in the same sentence, using exact numbers and units.}
  \end{itemize}
  \blk{Block 3}{the upper age limit (rare after 5, very rare after 7)}
  \begin{itemize}[nosep,leftmargin=1.2em,label=--]
    \crit{The response notes that the condition becomes significantly less common after age 5 and is very rare after age 7.}
    \crit{The response explains that incidence sharply declines after around age 5 due to ligament and bone development.}
    \crit{The response contains a distinct section titled ``Upper Age Limit'' that explicitly says the condition is ``significantly less common after age 5'' and ``very rare after age 7''.}
  \end{itemize}
  \blk{Block 4}{keeps the answer brief and focused on the age range}
  \begin{itemize}[nosep,leftmargin=1.2em,label=--]
    \crit{The response keeps the answer centered on the age range and does not include unrelated treatment or diagnosis details.}
    \crit{The response provides the age range in a brief, direct statement (e.g., first sentence) without unnecessary verbosity.}
  \end{itemize}
  \blk{Block 5}{explains the anatomical mechanism (ligament laxity, ossification)}
  \begin{itemize}[nosep,leftmargin=1.2em,label=--]
    \crit{The response adds a short explanation linking the risk to ligament laxity and skeletal development.}
    \crit{The response explains that the injury becomes unlikely in older children because the radial head ossifies/widens and the annular ligament strengthens.}
    \crit{The response includes a numbered list of three or more distinct anatomical factors (e.g., radial head shape, ligament laxity, ossification) that directly link skeletal maturation to the age-related risk.}
    \crit{The exact phrase ``ligament laxity'' is present, describing why younger children are predisposed to the injury.}
  \end{itemize}
\end{description}
\end{tcolorbox}
\caption{\textbf{Text-similarity blocks for a Medical rubric.} Same procedure as \cref{fig:block-example-sci}. All five blocks have two or more members, so every mask drops several criteria at once.}
\label{fig:block-example-med}
\end{figure*}

\paragraph{Leakage and effective dropout rate.}\clnote{Rewritten from the probe scripts; the 41/7 row is signal leakage over varying criteria with verdict-derived blocks. The 77/40 row and the derived ${\approx}30$ are from your summary and did not reproduce from the scripts.}
The left panel of \cref{tab:block} uses saved deep-plateau responses of the 8B Science base and $f{\approx}50\%$ runs to 8 ResearchQA evaluation prompts, 16 rollouts each, re-judged by the proxy judge. For two criteria of one rubric, $\phi$ is the correlation of their binary verdicts across the prompt's rollouts; it is defined only when both verdicts vary within the group, which are also the only criteria that carry GRPO signal. The \emph{leakage} of a mask is the mean, over its dropped varying criteria, of the largest $\phi$ (clipped at zero) to any surviving criterion. In the second row the blocks come from the verdicts themselves ($\phi\ge0.6$ components), not from the text blocks used in training, so the panel bounds what a correlation-aware mask can recover. The \emph{effective dropout rate} is $f_{\mathrm{eff}}=1-\rho^{2}$, with $\rho$ the correlation between masked and full-rubric reward across a group's rollouts: the fraction of reward variance the mask removes, which would equal $f$ if criteria were uncorrelated. It is measured for
independent masks by sampling masks on the logged verdicts, and derived for
block masks as $f\,(1-\text{leakage})$ from the first row.

\needspace{5\baselineskip}
\paragraph{Matched-overclaim comparison.}\clnote{These numbers come from the 600-step caches: Medical reproduces (28.5; 31.5 vs.\ 31.3), Science base 50.8 does not exactly (nearest 51.0--51.4).}
We compare gold at matched overclaim levels on the 600-step 8B runs, after the hacking onset. At these levels, the dropout runs match or exceed base on gold. On
Medical, at 40\% overclaim, base has 28.5\% gold and $f{\approx}50\%$ has
31.3\%. On Science, at 35\% overclaim, the numbers are 50.8\% versus 52.5\%.
Within each domain, at the same overclaim level, the dropout runs reach a higher gold score than base.

\paragraph{Comparison protocol.} Reward hacking grows with training, so end-of-run comparisons confound the effect with training length.
All cross-run comparisons use window means over steps 400--600 and matched-checkpoint win counts. The win counts record, at each eval checkpoint in the window shared by both runs, whether the run beats base on the identical prompt set. The window sits after the hacking onset (step $\sim$240 on Medical, \cref{fig:hacking}) and inside every run's logged range except that of the $f{\approx}60\%$ run, whose window statistics cover steps 400--540.
Win-count denominators differ where a run's logged evaluations end mid-window.

\paragraph{POW3R implementation.}
We implement POW3R \citep{tyagi2026pow3r} with the method's published defaults ($\lambda=0.5$, factors clipped to $[0.67, 1.5]$). Its dynamic per-criterion factors are derived from the within-group verdict variance of the current rollout group. The factors are applied once all rollouts in a group have been generated and judged, which is the same point at which GRPO forms advantages. We deviate from the
original in two ways.
RubricHub rubrics carry no per-criterion category labels, so the original's within-category balancing (their Eq.~3) is inactive here. As a result, the factors reweight criteria globally within each prompt's rubric.
We also recompute factors within each step from the current group's verdicts and apply them immediately, whereas the original smooths them across prompt visits with an EMA and a one-visit lag (their Eq.~7). Our implementation is therefore the unsmoothed variant.
Otherwise POW3R uses the same training recipe as every other run in this paper, and evaluation always uses the full static rubric.

\paragraph{Dropout variants not explored.}
Per-criterion fractions $f_k$ (with $f_k=0$ on criteria that must never be dropped) are one natural extension. Two others are schedules that anneal $f$ from high to zero over training, and blocks estimated online from the group's own verdicts instead of from criterion text. Viewed as an objective, dropout optimizes $\mathcal{J}_{\text{drop}}(\theta)=\mathbb{E}_m[\mathcal{J}(\theta;m)]$, a marginal over sub-rubrics, which formalizes ``never optimize the same rubric twice.''

\begin{figure}[tb]
  \centering
  \includegraphics[width=\linewidth]{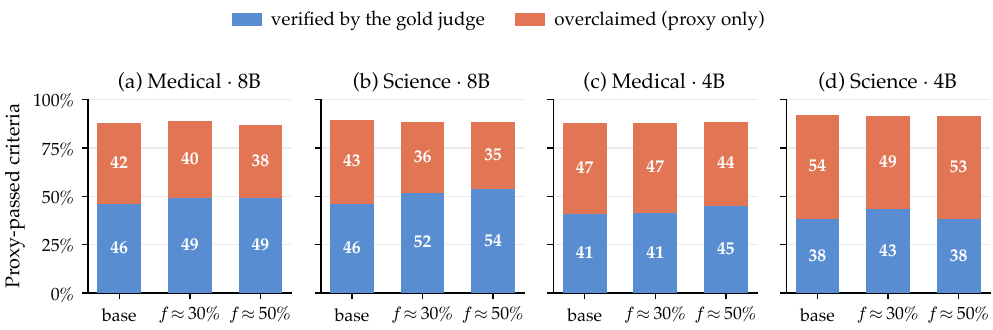}
  \caption{\textbf{Same proxy pass rate, different composition.}
Both judges grade each run's step-600 responses on matched prompts against every positive-weight criterion. Bar height is the share of criteria the proxy judge
  accepts, split into the part the gold judge confirms (blue) and the
  overclaimed part it rejects (terracotta).
\Cref{tab:mechanism} gives the
  full grid, with exact values, counts, and underclaim.}
  \label{fig:rejudge}
\end{figure}

\paragraph{Full per-criterion grid.}
\Cref{tab:mechanism} reports the complete step-600 per-criterion breakdown behind \cref{fig:rejudge} and adds the exact proxy pass rates and the underclaim row for both model sizes.

\input{tables/table_mechanism}

\clearpage
\section{Limitations}
\label{sec:limitations}
With one exception (Medical 4B, \cref{sec:setup}), every configuration is a single training run, so error bars reflect
within-run rather than across-seed variation. The gold judge is not ground
truth. Our claims rest on divergence and on run-to-run comparisons under
identical judges, which survive a constant judge bias but not a
distribution-dependent one. In-domain cost is measured on training prompts
only. The scope is one policy family at three sizes, two domains, and one RL
algorithm, and at 1.7B only the smaller fraction helps on Medical
(\cref{sec:main}). Block dropout is evaluated on both pairs at
three sizes and adds an offline clustering pass and a similarity threshold we did not sweep.

%% file: tables/table_ablation.tex
\begin{table}[tb]
  \centering
  \small
  \setlength{\tabcolsep}{8pt}
  \renewcommand{\arraystretch}{1.18}
  \begin{tabular}{lccccc}
    \toprule
    \textbf{Run} & Gold\,{\color{scaleGray}$\uparrow$} & $\Delta$\,{\color{scaleGray}$\uparrow$}
      & Proxy$-$gold\,{\color{scaleGray}$\downarrow$} & Overclaim\,{\color{scaleGray}$\downarrow$}
      & Train reward\,{\color{scaleGray}$\uparrow$} \\
    \midrule
    base            & $28.2$          & $\phantom{+}0.0$ & $40.3$ & $40.4$ & $\mathbf{98.0}$ \\
    \addlinespace[3pt]
    $f{\approx}20\%$      & $28.7$          & $+0.6$ & $40.2$ & $39.5$ & $97.7$ \\
    $f{\approx}30\%$      & $29.2$          & $+1.0$ & $38.7$ & $38.9$ & $97.8$ \\
    $f{\approx}40\%$      & $28.5$          & $+0.4$ & $41.3$ & $38.5$ & $97.6$ \\
    \rowcolor{scaleTeal!15}
    $f{\approx}50\%$      & $\mathbf{30.1}$ & $\mathbf{+2.0}$ & $\mathbf{38.4}$ & $\mathbf{37.2}$ & $97.6$ \\
    $f{\approx}60\%$      & $27.7$          & $-0.5$ & $42.6$ & $37.8$ & $97.3$ \\
    \midrule
    POW3R           & $27.0$          & $-1.2$ & $40.2$ & $42.2$ & $97.7$ \\
    \bottomrule
  \end{tabular}
  \caption{
\textbf{Ablations on the Medical pair} (window means over steps
  400--600, values in \%, $\Delta$ in points vs base). Bold marks the best value per column. The $f{\approx}50\%$ row is tinted with its figure color.}
  \label{tab:ablation}
\end{table}

%% file: tables/table_block.tex
\begin{table}[tb]
  \centering
  \begin{minipage}[t]{0.49\linewidth}
    \centering\scriptsize
    \setlength{\tabcolsep}{3pt}\renewcommand{\arraystretch}{1.05}
    \begin{tabular}[t]{lcc}
      \toprule
      \textbf{Offline, nominal $f{\approx}50\%$} & Independent & Block \\
      \midrule
      Leakage, all coupled pairs & 77 & 40 \\
      Leakage, varying criteria ($\phi{\ge}0.6$ blocks) & 41 & 7 \\
      Effective dropout rate $f_{\mathrm{eff}}$ & 20 & $\approx$30 \\
      \bottomrule
    \end{tabular}
  \end{minipage}\hfill
  \begin{minipage}[t]{0.49\linewidth}
    \centering\scriptsize
    \setlength{\tabcolsep}{3pt}\renewcommand{\arraystretch}{1.05}
    \begin{tabular}[t]{lcc}
      \toprule
      \textbf{Training, block $-$ independent} & $\Delta$ gold & $\Delta$ overclaim \\
      \midrule
      Science 8B   & $+1.5$ & $-1.7$ \\
      Science 4B   & $+1.8$ & $-2.4$ \\
      Science 1.7B & $+2.2$ & $-2.7$ \\
      Medical 8B   & $+0.8$ & $-2.7$ \\
      Medical 4B   & $+2.8$ & $-1.3$ \\
      Medical 1.7B & $+1.9$ & $-3.4$ \\
      \bottomrule
    \end{tabular}
  \end{minipage}
  \caption{\textbf{Block dropout: offline leak and training deltas.}
  Left: logged 8B Science verdicts at nominal $f{\approx}50\%$, in \%.
Leakage: mean over dropped criteria of the largest verdict correlation $\phi$ with a surviving criterion. The second row counts only criteria whose verdicts vary within the group. $f_{\mathrm{eff}}$: share of reward variance
  the mask removes, measured for independent masks and derived for block masks (\cref{sec:setup}).
  Right: block minus independent dropout, both at $f{\approx}50\%$, window means over steps 400--600 in
  points. Figure version in \cref{fig:blockstory}.}
  \label{tab:block}
\end{table}

%% file: tables/table_mechanism.tex
\begin{table}[!htb]
  \centering
  \small
  \setlength{\tabcolsep}{6pt}
  \renewcommand{\arraystretch}{1.12}
  \begin{tabular}{clccc@{\hspace{10pt}}ccc}
    \toprule
    & & \multicolumn{3}{c}{\textbf{Medical}} & \multicolumn{3}{c}{\textbf{Science}} \\
    \cmidrule(lr){3-5}\cmidrule(l){6-8}
    & \textbf{Matched step 600} & base & $f{\approx}30\%$ & $f{\approx}50\%$ & base & $f{\approx}30\%$ & $f{\approx}50\%$ \\
    \midrule
    \multirow{4}{*}{\emph{8B}}
    & proxy pass rate                 & $87.7$ & $88.8$ & $87.1$ & $89.5$ & $88.3$ & $88.3$ \\
    & gold pass rate ($\uparrow$)     & $48.7$ & $51.5$ & $\mathbf{52.3}$ & $46.0$ & $51.5$ & $\mathbf{53.4}$ \\
    & overclaim ($\downarrow$)        & $41.5$ & $39.7$ & $\mathbf{37.9}$ & $43.4$ & $36.4$ & $\mathbf{34.7}$ \\
    & underclaim ($\downarrow$)       & $2.5$  & $\mathbf{2.4}$  & $3.1$ & $0.1$ & $0.1$ & $0.1$ \\
    \midrule
    \multirow{4}{*}{\emph{4B}}
    & proxy pass rate                 & $87.7$ & $88.0$ & $88.5$ & $92.2$ & $91.8$ & $91.5$ \\
    & gold pass rate ($\uparrow$)     & $43.4$ & $43.7$ & $\mathbf{47.3}$ & $38.1$ & $\mathbf{43.3}$ & $38.2$ \\
    & overclaim ($\downarrow$)        & $46.9$ & $46.8$ & $\mathbf{43.7}$ & $54.1$ & $\mathbf{48.6}$ & $53.4$ \\
    & underclaim ($\downarrow$)       & $2.6$  & $\mathbf{2.4}$  & $2.5$ & $\mathbf{0.0}$ & $0.1$ & $0.1$ \\
    \bottomrule
  \end{tabular}
  \caption{
\textbf{Per-criterion breakdown at step 600 for Medical and Science, 8B and 4B} (all values in \%). The table gives all values underlying \cref{fig:rejudge}.
  Both judges grade every positive-weight criterion of each run's step-600
  responses on matched prompts (temp.\ 0).
8B: Medical 992 prompts, 7{,}608 criteria; Science 362--363
  prompts. 4B: Medical 992 prompts, 7{,}667 criteria; Science 368 prompts. Overclaim: proxy accepts, gold rejects. Underclaim:
  proxy rejects, gold accepts. Bold: best run per row, pair, and block.}
  \label{tab:mechanism}
\end{table}

%% file: references.bib
@article{amodei2016concrete,
  title   = {Concrete Problems in {AI} Safety},
  author  = {Amodei, Dario and Olah, Chris and Steinhardt, Jacob and Christiano, Paul and Schulman, John and Man{\'e}, Dan},
  journal = {arXiv preprint arXiv:1606.06565},
  year    = {2016}
}

@inproceedings{skalse2022reward,
  title     = {Defining and Characterizing Reward Gaming},
  author    = {Skalse, Joar and Howe, Nikolaus H. R. and Krasheninnikov, Dmitrii and Krueger, David},
  booktitle = {Advances in Neural Information Processing Systems (NeurIPS)},
  volume    = {35},
  year      = {2022}
}

@inproceedings{pan2022misspecification,
  title     = {The Effects of Reward Misspecification: Mapping and Mitigating Misaligned Models},
  author    = {Pan, Alexander and Bhatia, Kush and Steinhardt, Jacob},
  booktitle = {International Conference on Learning Representations (ICLR)},
  year      = {2022}
}

@inproceedings{gao2023overoptimization,
  title     = {Scaling Laws for Reward Model Overoptimization},
  author    = {Gao, Leo and Schulman, John and Hilton, Jacob},
  booktitle = {Proceedings of the 40th International Conference on Machine Learning (ICML)},
  series    = {Proceedings of Machine Learning Research},
  volume    = {202},
  pages     = {10835--10866},
  year      = {2023},
  publisher = {PMLR}
}

@inproceedings{zheng2023mtbench,
  title     = {Judging {LLM}-as-a-Judge with {MT-Bench} and Chatbot Arena},
  author    = {Zheng, Lianmin and Chiang, Wei-Lin and Sheng, Ying and Zhuang, Siyuan and Wu, Zhanghao and Zhuang, Yonghao and Lin, Zi and Li, Zhuohan and Li, Dacheng and Xing, Eric P. and Zhang, Hao and Gonzalez, Joseph E. and Stoica, Ion},
  booktitle = {Advances in Neural Information Processing Systems (NeurIPS) Datasets and Benchmarks Track},
  volume    = {36},
  year      = {2023}
}

@article{shao2024deepseekmath,
  title   = {{DeepSeekMath}: Pushing the Limits of Mathematical Reasoning in Open Language Models},
  author  = {Shao, Zhihong and Wang, Peiyi and Zhu, Qihao and Xu, Runxin and Song, Junxiao and Bi, Xiao and Zhang, Haowei and Zhang, Mingchuan and Li, Y. K. and Wu, Y. and Guo, Daya},
  journal = {arXiv preprint arXiv:2402.03300},
  year    = {2024}
}

@article{srivastava2014dropout,
  title   = {Dropout: A Simple Way to Prevent Neural Networks from Overfitting},
  author  = {Srivastava, Nitish and Hinton, Geoffrey and Krizhevsky, Alex and Sutskever, Ilya and Salakhutdinov, Ruslan},
  journal = {Journal of Machine Learning Research},
  volume  = {15},
  number  = {56},
  pages   = {1929--1958},
  year    = {2014}
}

@inproceedings{lambert2024tulu3,
  title     = {{T{\"u}lu 3}: Pushing Frontiers in Open Language Model Post-Training},
  author    = {Lambert, Nathan and Morrison, Jacob and Pyatkin, Valentina and Huang, Shengyi and Ivison, Hamish and Brahman, Faeze and Miranda, Lester James Validad and Liu, Alisa and Dziri, Nouha and Lyu, Xinxi and Gu, Yuling and Malik, Saumya and others},
  booktitle = {Conference on Language Modeling (COLM)},
  year      = {2025}
}

@article{arora2025healthbench,
  title   = {{HealthBench}: Evaluating Large Language Models Towards Improved Human Health},
  author  = {Arora, Rahul K. and Wei, Jason and Soskin Hicks, Rebecca and Bowman, Preston and Qui{\~n}onero-Candela, Joaquin and Tsimpourlas, Foivos and Sharman, Michael and Shah, Meghan and Vallone, Andrea and Beutel, Alex and Heidecke, Johannes and Singhal, Karan},
  journal = {arXiv preprint arXiv:2505.08775},
  year    = {2025}
}

@article{qwen2025qwen3,
  title   = {{Qwen3} Technical Report},
  author  = {Yang, An and Li, Anfeng and Yang, Baosong and Zhang, Beichen and Hui, Binyuan and Zheng, Bo and Yu, Bowen and Gao, Chang and Huang, Chengen and Lv, Chenxu and Zheng, Chujie and Liu, Dayiheng and others},
  journal = {arXiv preprint arXiv:2505.09388},
  year    = {2025}
}

@inproceedings{gunjal2025rubrics,
  title     = {Rubrics as Rewards: Reinforcement Learning Beyond Verifiable Domains},
  author    = {Gunjal, Anisha and Wang, Anthony and Lau, Elaine and Nath, Vaskar and He, Yunzhong and Liu, Bing and Hendryx, Sean},
  booktitle = {International Conference on Learning Representations (ICLR)},
  pages     = {127924--127945},
  year      = {2026}
}

@article{huang2025rubric,
  title   = {Reinforcement Learning with Rubric Anchors},
  author  = {Huang, Zenan and Zhuang, Yihong and Lu, Guoshan and Qin, Zeyu and Xu, Haokai and Zhao, Tianyu and Peng, Ru and Hu, Jiaqi and Shen, Zhanming and Hu, Xiaomeng and others},
  journal = {arXiv preprint arXiv:2508.12790},
  year    = {2025}
}

@article{tyagi2026pow3r,
  title   = {Not Every Rubric Teaches Equally: Policy-Aware Rubric Rewards for {RLVR}},
  author  = {Tyagi, Utkarsh and Guo, Xingang and Rezaei, MohammadHossein and George, Daniel and Mahmoud, Anas and Lee, Jackson and Liu, Bing and He, Yunzhong},
  journal = {arXiv preprint arXiv:2605.20164},
  year    = {2026}
}

@inproceedings{liu2026gdpo,
  title   = {{GDPO}: Group reward-Decoupled Normalization Policy Optimization for Multi-reward {RL} Optimization},
  author  = {Liu, Shih-Yang and Dong, Xin and Lu, Ximing and Diao, Shizhe and Belcak, Peter and Liu, Mingjie and Chen, Min-Hung and Yin, Hongxu and Wang, Yu-Chiang Frank and Cheng, Kwang-Ting and Choi, Yejin and Kautz, Jan and Molchanov, Pavlo},
  booktitle = {Proceedings of the 43rd International Conference on Machine Learning ({ICML})},
  year    = {2026}
}

@article{deepseekai2025r1,
  title   = {{DeepSeek-R1} incentivizes reasoning in {LLMs} through reinforcement learning},
  author  = {Guo, Daya and Yang, Dejian and Zhang, Haowei and Song, Junxiao and Wang, Peiyi and Zhu, Qihao and Xu, Runxin and Zhang, Ruoyu and Ma, Shirong and Bi, Xiao and Zhang, Xiaokang and Yu, Xingkai and others},
  journal = {Nature},
  volume  = {645},
  number  = {8081},
  pages   = {633--638},
  year    = {2025},
  doi     = {10.1038/s41586-025-09422-z}
}

@article{wang2026cherrl,
  title   = {Reproducing, Analyzing, and Detecting Reward Hacking in Rubric-Based Reinforcement Learning},
  author  = {Wang, Xuekang and Hao, Zhuoyuan and Hou, Shuo and Peng, Hao and Li, Juanzi and Wang, Xiaozhi},
  journal = {arXiv preprint arXiv:2606.04923},
  year    = {2026}
}

@article{mahmoud2026rewardhacking,
  title   = {Reward Hacking in Rubric-Based Reinforcement Learning},
  author  = {Mahmoud, Anas and Rezaei, MohammadHossein and Wang, Zihao and Gunjal, Anisha and Liu, Bing and He, Yunzhong},
  journal = {arXiv preprint arXiv:2605.12474},
  year    = {2026}
}

@article{researchqa2025,
  title   = {{ResearchQA}: Evaluating Scholarly Question Answering at Scale Across 75 Fields with Survey-Mined Questions and Rubrics},
  author  = {Yifei, Li S. and Chang, Allen and Malaviya, Chaitanya and Yatskar, Mark},
  journal = {Transactions of the Association for Computational Linguistics},
  volume  = {14},
  pages   = {1344--1368},
  year    = {2026},
  doi     = {10.1162/TACL.a.732}
}

@inproceedings{advancedif,
  title     = {{A}dvanced{IF}: Rubric-Based Benchmarking and Reinforcement Learning for Advancing {LLM} Instruction Following},
  author    = {He, Yun and Li, Wenzhe and Zhang, Hejia and Li, Songlin and Mandyam, Karishma and Khosla, Sopan and Xiong, Yuanhao and Wang, Nanshu and Peng, Xiaoliang and Li, Beibin and Bi, Shengjie and Patil, Shishir G and others},
  booktitle = {Proceedings of the 64th Annual Meeting of the Association for Computational Linguistics (Volume 1: Long Papers)},
  year      = {2026},
  month     = jul,
  pages     = {18003--18022},
  address   = {San Diego, California, United States},
  publisher = {Association for Computational Linguistics},
  doi       = {10.18653/v1/2026.acl-long.820},
  url       = {https://aclanthology.org/2026.acl-long.820/}
}

@inproceedings{openrubrics,
  title = {{OpenRubrics}: Towards Scalable Synthetic Rubric Generation for Reward Modeling and {LLM} Alignment},
  author = {Liu, Tianci and Xu, Ran and Yu, Tony and Hong, Ilgee and Yang, Carl and Zhao, Tuo and Wang, Haoyu},
  booktitle = {Proceedings of the 64th Annual Meeting of the Association for Computational Linguistics (Volume 1: Long Papers)},
  pages = {17417--17437},
  publisher = {Association for Computational Linguistics},
  address = {San Diego, California, United States},
  doi = {10.18653/v1/2026.acl-long.791},
  year = {2026}
}

@inproceedings{checklists,
  title={Checklists Are Better Than Reward Models For Aligning Language Models},
  author={Viswanathan, Vijay and Sun, Yanchao and Kong, Xiang and Cao, Meng and Neubig, Graham and Wu, Tongshuang},
  booktitle={Advances in Neural Information Processing Systems ({NeurIPS})},
  volume={38},
  pages={127195--127221},
  year={2025}
}

@inproceedings{ruscarl,
  title = {Breaking the Exploration Bottleneck: {R}ubric-Scaffolded Reinforcement Learning for Open-Ended {LLM} Reasoning},
  author = {Zhou, Yang and Li, Sunzhu and Liu, Shunyu and Fang, Wenkai and Zhang, Kongcheng and Zhao, Jiale and Yang, Jingwen and Zhou, Yihe and Lv, Jianwei and Zheng, Tongya and Lu, Hengtong and Chen, Wei and others},
  booktitle = {Proceedings of the 43rd International Conference on Machine Learning ({ICML})},
  year = {2026}
}

@inproceedings{coste2024ensembles,
  title={Reward Model Ensembles Help Mitigate Overoptimization},
  author={Coste, Thomas and Anwar, Usman and Kirk, Robert and Krueger, David},
  booktitle={The Twelfth International Conference on Learning Representations ({ICLR})},
  year={2024}
}

@inproceedings{rame2024warm,
  title     = {{WARM}: On the Benefits of Weight Averaged Reward Models},
  author    = {Rame, Alexandre and Vieillard, Nino and Hussenot, Leonard and Dadashi-Tazehozi, Robert and Cideron, Geoffrey and Bachem, Olivier and Ferret, Johan},
  booktitle = {Proceedings of the 41st International Conference on Machine Learning},
  series    = {Proceedings of Machine Learning Research},
  volume    = {235},
  pages     = {42048--42073},
  publisher = {PMLR},
  year      = {2024}
}

@inproceedings{eisenstein2024herding,
  title     = {Helping or Herding? {R}eward Model Ensembles Mitigate but do not Eliminate Reward Hacking},
  author    = {Eisenstein, Jacob and Nagpal, Chirag and Agarwal, Alekh and Beirami, Ahmad and D'Amour, Alex and Dvijotham, {DJ} and Fisch, Adam and Heller, Katherine and Pfohl, Stephen and Ramachandran, Deepak and Shaw, Peter and Berant, Jonathan},
  booktitle = {First Conference on Language Modeling ({COLM})},
  year      = {2024},
  url       = {https://openreview.net/forum?id=5u1GpUkKtG}
}

@inproceedings{chen2024odin,
  title     = {{ODIN}: Disentangled Reward Mitigates Hacking in {RLHF}},
  author    = {Chen, Lichang and Zhu, Chen and Chen, Jiuhai and Soselia, Davit and Zhou, Tianyi and Goldstein, Tom and Huang, Heng and Shoeybi, Mohammad and Catanzaro, Bryan},
  booktitle = {Proceedings of the 41st International Conference on Machine Learning},
  series    = {Proceedings of Machine Learning Research},
  volume    = {235},
  pages     = {7935--7952},
  publisher = {PMLR},
  year      = {2024}
}

@inproceedings{zhang-etal-2025-llmeval,
    title = "{LLME}val-{M}ed: A Real-world Clinical Benchmark for Medical {LLM}s with Physician Validation",
    author = "Zhang, Ming  and
      Shen, Yujiong  and
      Li, Zelin  and
      Sha, Huayu  and
      Hu, Binze  and
      Wang, Yuhui  and
      Huang, Chenhao  and
      Liu, Shichun  and
      Tong, Jingqi  and
      Jiang, Changhao  and
      Chai, Mingxu  and
      Xi, Zhiheng  and
      Dou, Shihan  and
      Gui, Tao  and
      Zhang, Qi  and
      Huang, Xuanjing",
    editor = "Christodoulopoulos, Christos  and
      Chakraborty, Tanmoy  and
      Rose, Carolyn  and
      Peng, Violet",
    booktitle = "Findings of the Association for Computational Linguistics: EMNLP 2025",
    month = nov,
    year = "2025",
    address = "Suzhou, China",
    publisher = "Association for Computational Linguistics",
    url = "https://aclanthology.org/2025.findings-emnlp.263/",
    doi = "10.18653/v1/2025.findings-emnlp.263",
    pages = "4888--4914",
    ISBN = "979-8-89176-335-7"
}

@inproceedings{dineen2025qalign,
  title={{QA-LIGN}: Aligning {LLMs} through Constitutionally Decomposed {QA}},
  author={Dineen, Jacob and Rrv, Aswin and Liu, Qin and Xu, Zhikun and Ye, Xiao and Shen, Ming and Li, Zhaonan and Lu, Shijie and Baral, Chitta and Chen, Muhao and Zhou, Ben},
  booktitle={Findings of the Association for Computational Linguistics: EMNLP 2025},
  pages={20619--20642},
  year={2025}
}

@inproceedings{dineen2026vocabdropout,
  title={Vocabulary Dropout for Curriculum Diversity in {LLM} Co-Evolution},
  author={Dineen, Jacob and RRV, Aswin and Xu, Zhikun and Zhou, Ben},
  booktitle={Conference on Language Modeling ({COLM})},
  year={2026}
}

@article{li2026rubrichub,
  title   = {RubricHub: A Comprehensive and Highly Discriminative Rubric Dataset via Automated Coarse-to-Fine Generation},
  author  = {Li, Sunzhu and Zhao, Jiale and Wei, Miteto and Ren, Huimin and Zhou, Yang and Yang, Jingwen and Liu, Shunyu and Zhang, Kaike and Chen, Wei},
  journal = {arXiv preprint arXiv:2601.08430},
  year    = {2026},
  url     = {https://arxiv.org/abs/2601.08430}
}
